\documentclass{article} %
\usepackage{iclr2027_conference,times}

\usepackage{amsmath,amsfonts,bm}

\def\eqref#1{equation~\ref{#1}}

\def\1{\bm{1}}

\DeclareMathAlphabet{\mathsfit}{\encodingdefault}{\sfdefault}{m}{sl}
\SetMathAlphabet{\mathsfit}{bold}{\encodingdefault}{\sfdefault}{bx}{n}

\usepackage{multirow}
\usepackage{siunitx}
\usepackage{hyperref}
\usepackage{url}
\usepackage{booktabs} %

\title{ACTS: Learning Visuo-Tactile World Models from Human Physical Interaction}

\author{%
Yuxiang Ma$^{1}$\thanks{Equal contribution.} \quad
Weihan Xu$^{2}$\footnotemark[1] \quad
Pengfei Ye$^{1}$ \quad
Anna Mei$^{3}$ \quad
Yuxin Ray Song$^{1}$ \\
\textbf{Tingle Li$^{4}$ \quad
Haonan Chen$^{5}$ \quad
Edward H.~Adelson$^{1}$} \\[4pt]
{\normalfont $^{1}$Massachusetts Institute of Technology \quad
$^{2}$University of Washington \quad
$^{3}$Harvard University} \\
{\normalfont $^{4}$University of California, Berkeley \quad
$^{5}$Stanford University}
}

\usepackage{graphicx}

\iclrfinalcopy
\begin{document}

\maketitle
\suppressfloats[t] %

\begin{abstract}
Action-conditioned world models for contact-rich manipulation require data that captures how contact forms, slides, and releases under varying force. Such data are slow to collect through robot teleoperation, which often lacks force feedback and is constrained by gripper geometry. We ask whether these dynamics can instead be learned from human physical interaction and transferred to a robot.
We present ACTS, a visuo-tactile world model trained only on human interaction data. Operators manipulate handheld GelSight sensors while we record third-person and wrist video, tactile images, 6-DoF sensor motion, and tactile-derived normal force, yielding React, a 4.9-hour bimanual dataset across three objects. We represent actions at the tactile sensor as motion augmented by force change, providing a shared action space for human collection and robot execution. ACTS jointly predicts future scene, wrist, and tactile observations.
On held-out human data, ACTS improves scene-view prediction by 0.7--2.0\,dB PSNR over the stronger of two visuo-tactile world models trained from scratch on React. Without fine-tuning, in the robot teleoperation rollouts we inspected, predicted tactile imprints still appear and disappear with recorded contact. Visual transfer is less robust: camera predictions exceed frame copying on PushT, are mixed on Rope, and fall below it on Motherboard, while wrist views can blur or hallucinate structure. These results suggest that contact dynamics transfer more reliably than visual appearance from human to robot.
Project website and supplementary videos: \url{https://acts-anon.github.io/}.
\end{abstract}

\section{Introduction}

\begin{figure}[t]
    \vspace{-10pt}
    \centering
    \includegraphics[width=\textwidth]{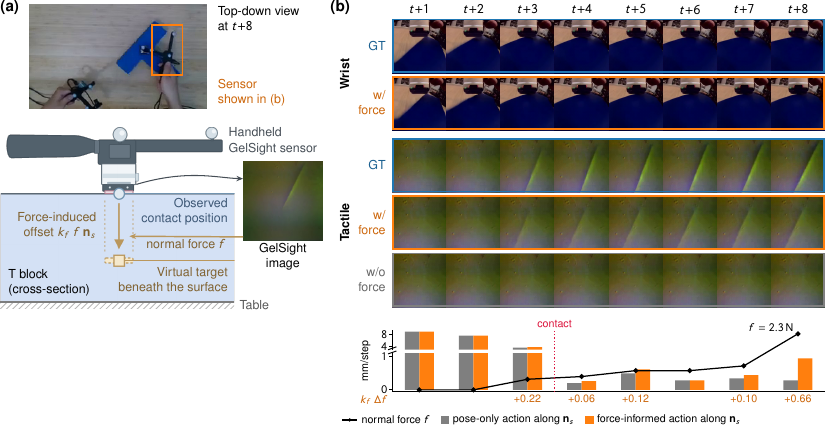}
    \caption{\textbf{Force-informed sensor actions.}
    (a)~Top: scene view, with the sensor of (b) boxed. Estimated normal force $f$ shifts the sensor target by
    $k_f f\,\mathbf{n}_s$ along the gel normal, so each action includes
    $k_f\Delta f\,\mathbf{n}_s$ (Eq.~\ref{eq:sensor_action}).
    (b)~On a held-out PushT sequence, the force-conditioned model forms the
    tactile imprint at contact, while the pose-only model does not. Tactile
    crops are shown at 2$\times$ contrast.}
    \label{fig:teaser}
    \vspace{-12pt}
\end{figure}

Robot world models predict how candidate actions change future observations,
supporting planning and policy learning~\citep{zhu2025irasim,hou2026worldmodelsurvey}.
Contact-rich manipulation is difficult to model because small changes in
contact location and force can determine whether an object sticks, slides,
jams, or loses contact. Vision captures scene motion but often cannot resolve
the local interactions that cause it. Learning such dynamics therefore
requires data spanning contact formation, loading, sliding, and release under
varied geometry and force. Collecting this interaction data is the bottleneck
we address.

Existing visuo-tactile world models are trained on robot interaction
trajectories~\citep{higuera2026visuotactileworldmodels,huang2026vitacworld},
in one case combined with data from a handheld gripper~\citep{zheng2026omnivta}.
In parallel, robot-free interfaces such as UMI and
TacUMI collect human demonstrations with handheld devices
~\citep{chi2024universal,cheng2026tacumi}. Robot teleoperation is slow for
dense contact exploration and typically provides limited force
feedback~\citep{tang2026towards},
while handheld grippers remove the robot but retain a fixed contact geometry.
In both cases, actions are tied to the collection device. We instead define
motion and contact at the tactile sensor, where the same action representation
can be used by a human or a robot.

Human operators directly manipulate handheld GelSight sensors while we record
6-DoF sensor poses together with synchronized visual and tactile observations.
We represent each action by relative sensor motion augmented with a
force-informed displacement~\citep{chen2025dexforceextractingforceinformedactions}
along the sensor normal, derived from tactile-estimated normal force
(Fig.~\ref{fig:teaser}). This interface allows operators to vary contact
location, orientation, and loading without a gripper constraining how the
sensor meets the object. Because the action is defined at the sensor rather
than the hand that carries it, human collection and robot execution share the
same sensor-level action space.

We use these interactions to train \textbf{ACTS}, a visuo-tactile world model
adapted from a pretrained video model. Given observation history and sensor
actions, ACTS jointly predicts future scene, wrist, and tactile observations.
Each action is provided both as force-informed motion in the local sensor frame
and as sensor geometry projected into the scene view. We collect
\textbf{React}, 4.9 hours of curated bimanual interaction with three objects,
including multi-view video, two GelSight Mini streams
~\citep{johnson2009retrographic,yuan2017gelsight}, tracked sensor poses, and
tactile-derived normal force.

We train one model per object using only human data and evaluate zero-shot on
robot teleoperation with the same sensing devices. ACTS outperforms
visuo-tactile world-model baselines on held-out human scene prediction, while
robot rollouts show that predicted tactile contact transitions can transfer
without robot-specific fine-tuning. Visual transfer is less robust, revealing
a remaining gap between human and robot appearance distributions.

Our contributions are:
{\setlength{\leftmargini}{1.2em}
\begin{itemize}
    \setlength{\itemsep}{0pt}\setlength{\parsep}{0pt}
    \item A sensor-level action representation that provides a common interface
    for human data collection and robot execution.
    \item \textbf{React}, a 4.9-hour human visuo-tactile dataset with tracked
    sensor poses and tactile-derived normal force.\footnote{We will release the
    React dataset, trained models, and code upon acceptance.}
    \item \textbf{ACTS}, a visuo-tactile world model trained only on human
    interaction data and evaluated zero-shot on robot teleoperation.
\end{itemize}}
\section{Related Work}

\noindent\textbf{World models for robot manipulation.}
World models learn action-conditioned dynamics for planning, simulation, policy learning, and evaluation~\citep{hou2026worldmodelsurvey,wang2026worldmodelsroboticmanipulation}. Visual-foresight methods planned through predicted image trajectories~\citep{finn2017deep,dasari2020robonet}, and recent models build on pretrained visual representations and video generators~\citep{zhou2025dinowm,zhou2024robodreamer,zhu2025irasim,guo2025ctrl}. These models learn from robot trajectories, with actions given as end-effector motion, joint commands, or gripper states~\citep{dasari2020robonet,zhu2025irasim,guo2025ctrl}, tying the learned dynamics to the robot that collected the data.

\noindent\textbf{Visuo-tactile world models.}
Tactile sensing resolves contact that vision leaves ambiguous~\citep{shan2026learningphysicalinteractionsurvey}, and several world models now predict tactile futures. VT-WM~\citep{higuera2026visuotactileworldmodels} jointly predicts visual and tactile latents for planning, ViTacWorld~\citep{huang2026vitacworld} trains on combined real and simulated data, and OmniVTA~\citep{zheng2026omnivta} pairs short-horizon visuo-tactile prediction with a contact-aware policy. ContactWorld~\citep{zhang2026contactworld} compares visual and tactile representations for predictive planning, FeelWorld~\citep{ma2026feelworld} predicts contact state, a force-related tactile latent, and slip, and Dream-Tac~\citep{lou2026dreamtac} and VT-WAM~\citep{tian2026vtwam} couple tactile prediction with action generation. These models learn from robot trajectories; OmniVTA adds demonstrations from a handheld gripper~\citep{cheng2026tacumi}, so its actions remain those of a gripper. ACTS learns from human interaction with bare sensors instead, with actions defined at the tactile sensor so that the same representation describes a robot moving that sensor.

\noindent\textbf{Visuo-tactile datasets.}
Visuo-tactile datasets record or simulate objects~\citep{gao2022objectfolder}, human touch in natural scenes~\citep{yang2022touch}, spatially aligned visual--tactile samples~\citep{dou2024tarf}, in-hand manipulation~\citep{suresh2024neuralfeels}, and data pooled across sensors~\citep{zhao2025t3}. They broaden tactile coverage but generally do not provide the action supervision a world model needs. Handheld collection systems do record actions: UMI~\citep{chi2024universal} introduced robot-free demonstration collection, FreeTacMan, exUMI, and TacUMI add tactile, force, and richer motion sensing~\citep{wu2026freetacman,xu2025exumi,cheng2026tacumi}, and MimicTouch~\citep{yu2025mimictouch} records tactile demonstrations from the human hand. These systems target policy learning. React instead pairs visual and tactile observations with sensor-level motion and normal force, the supervision needed to train an action-conditioned world model.

\section{Dataset}

\noindent\textbf{Human interaction data.}
React is collected by five operators performing bimanual interactions with a motherboard, a T-shaped block (PushT), and a rope. Each operator holds two independently movable devices, each equipped with a GelSight Mini vision-based tactile sensor~\citep{yuan2017gelsight}, an OptiTrack marker assembly for 6-DoF pose tracking, and a local camera. Three fixed cameras record the workspace (Fig.~\ref{fig:data_collection_setup}, Appendix~\ref{app:dataset}).
Operators slide, pivot, push, and pull the objects while varying contact location, orientation, and loading. The data cover making and breaking contact, sticking, sliding, and object motion. Operators explore these contact conditions without pursuing a prescribed task goal. After curation, the human dataset contains 4.91~hours of interaction across 187 segments from 60 source recordings.

\noindent\textbf{Robot evaluation data.}\label{robot_data} 
We collect a separate bimanual evaluation set with two YAM arms controlled through GELLO teleoperation interfaces~\citep{wu2024gello}. Each arm carries the same handheld device used for human collection, with its GelSight Mini, marker assembly, and wrist camera, so robot and human episodes share sensors and action construction. Teleoperators perform non-prehensile manipulation with the same three objects. We record five one-minute episodes per object, yielding 15 episodes and 15 minutes of robot interaction. The world model is trained only on human interaction data; all robot episodes are held out for evaluation.

\noindent\textbf{Post-processing.} We align the visual, tactile, and pose streams by timestamp to a common 30~Hz timeline and retain flags that distinguish new tactile captures from repeated frames. We remove intervals with sensor freezes or tracking failures and divide the remaining data into contiguous valid segments. Normal-force estimates derived from the tactile images are combined with the tracked sensor motion to construct force-informed actions. Appendix~\ref{app:dataset} gives the alignment procedure, force-estimation method, and dataset statistics.

\vspace{-2ex}
\section{Method}
\label{sec:method}

\begin{figure} [t]
\vspace{-10 pt}
    \centering
    \includegraphics[width=0.9\linewidth]{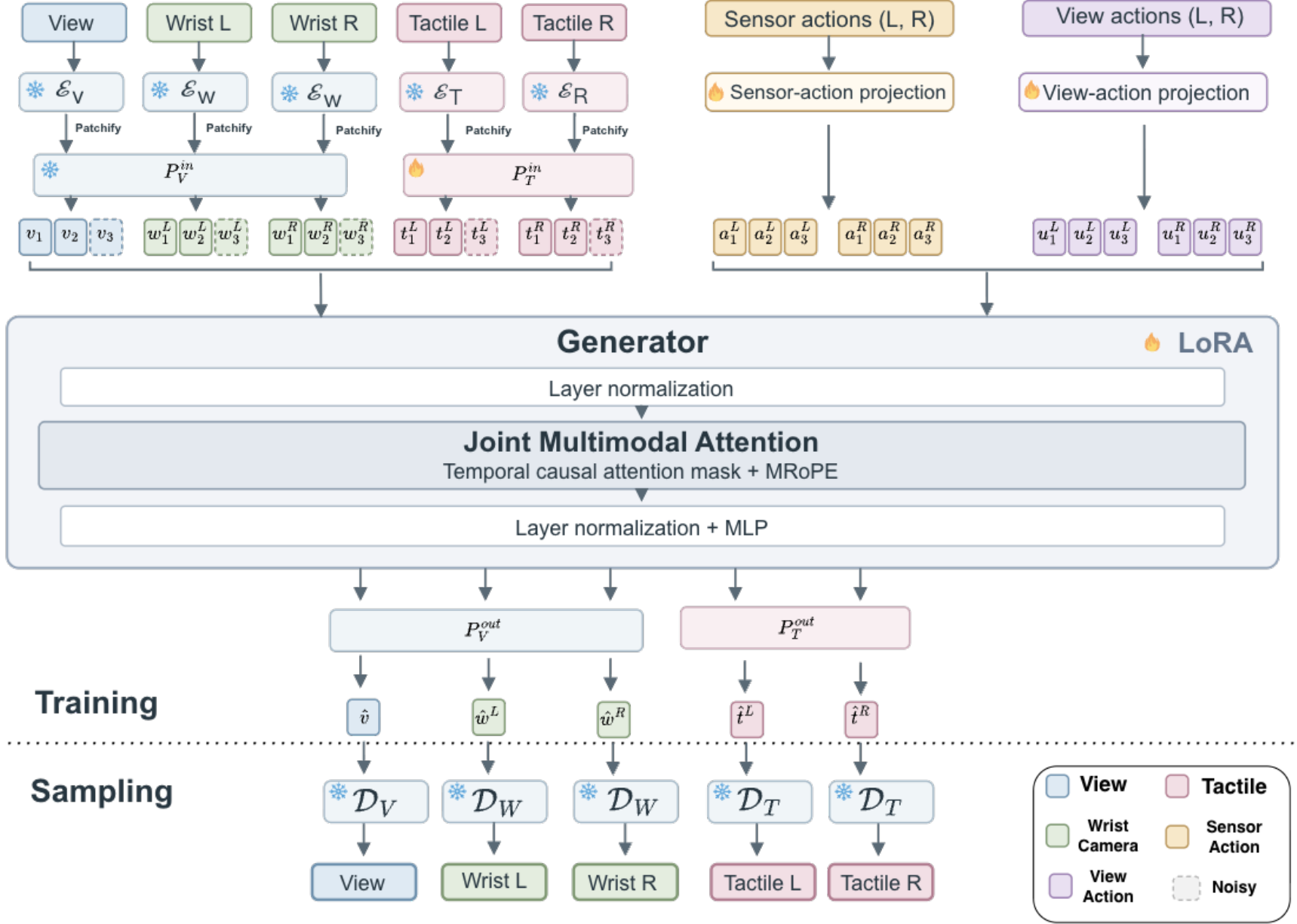}
\caption{
\textbf{Overview of ACTS.}
Frozen encoders produce observation latents, whose spatial
patches are mapped to tokens by input projectors $P^{\mathrm{in}}$.
A generator initialized from pretrained Cosmos3 weights
jointly processes noisy future tokens conditioned on clean
history, force-aware sensor actions, and view-aligned actions.
Output projectors $P^{\mathrm{out}}$ map target-token features
to latent velocity predictions.
The scene and two wrist streams share the RGB projector pair
$(P_V^{\mathrm{in}},P_V^{\mathrm{out}})$, while both tactile
streams share $(P_T^{\mathrm{in}},P_T^{\mathrm{out}})$.
Sampling updates only future latents, which frozen decoders
reconstruct into synchronized multimodal predictions.
}
    \label{fig:method}
    \vspace{-10 pt}
\end{figure}

We propose ACTS (Fig.~\ref{fig:method}), an action-conditioned world model jointly
predicting future scene video, wrist views, and tactile observations
from observation histories and actions for bimanual manipulation.
We describe observation encoding and tokenization
(Sec.~\ref{sec:observation_representation}), then force-aware
sensor actions and view-aligned actions encoding contact-related
motion and scene-view sensor geometry
(Sec.~\ref{sec:action_conditioning}).
We detail unified sequence construction with spatiotemporal
alignment and full-window action conditioning
(Sec.~\ref{sec:multimodal_transformer}), alongside the
training objective and inference protocol
(Sec.~\ref{sec:training_and_rollout}).

\subsection{Multimodal Observation Representation}
\label{sec:observation_representation}

To capture manipulation from complementary perspectives, we jointly
model scene video, wrist views, and tactile observations in a shared
latent prediction framework.
We define the observation streams as
$\mathcal{M}=\{V,W^L,W^R,T^L,T^R\}$, where $V$ denotes the
top-down scene RGB view, $W^L$ and $W^R$ denote the left and
right wrist-camera RGB views, and $T^L$ and $T^R$ denote
the corresponding GelSight observations.
For each stream $m\in\mathcal{M}$, let $X_H^m$ and $X_F^m$
denote its observed history and future sequence, respectively.
Given synchronized histories and action conditions $\mathcal{A}$,
our goal is to model
\begin{equation}
    p_\theta\!\left(
        \{X_F^m\}_{m\in\mathcal{M}}
        \mid \{X_H^m\}_{m\in\mathcal{M}},\mathcal{A}
    \right),
    \label{eq:acts_prediction}
\end{equation}
where $\theta$ denotes the model's trainable parameters.

To obtain compact representations while preserving temporal
causality, we encode each observation stream as
\begin{equation}
    Z^m = \mathcal{E}_m(X^m),
    \qquad m\in\mathcal{M},
    \label{eq:acts_encoding}
\end{equation}
where $X^m$ concatenates history $X_H^m$ and future $X_F^m$
in temporal order.
The scene and two wrist-camera streams share the pretrained
Wan2.2 causal VAE encoder~\citep{wan2025}, while both GelSight
streams share a variant adapted to our visuo-tactile
dataset.\footnote{Details of tactile VAE adaptation are provided
in Appendix~\ref{sec:tactile_encoder}.}
All encoders are frozen, and observations are encoded offline.
We partition $Z^m$ into history latents $Z_H^m$ and future
latent blocks $Z_F^m=(z_1^m,\ldots,z_Q^m)$.
We spatially patchify each latent block and linearly project
the patches into a common $d$-dimensional token space,
adding learned stream embeddings to distinguish their sources.
Let $\mathbf{e}_{q,p}^{m}\in\mathbb{R}^{d}$ denote the token
for patch $p$ of future block $q$ in stream $m$;
history blocks are tokenized identically.
The scene and two wrist-camera streams share one input
projection and one output projection, while the two GelSight
streams share a separate pair.
The output projections map hidden states back to the
corresponding latent patch spaces.

\subsection{Action Representation}
\label{sec:action_conditioning}
We connect motion, contact dynamics, and scene geometry through
two complementary action representations.
Following the trajectory-conditioned setting of IRASim~\citep{zhu2025irasim} and
Cosmos3~\citep{nvidia2026cosmos3omnimodalworld}, we provide
both representations for both hands over the complete window,
collectively denoted by $\mathcal{A}$, as clean conditions
to every predicted observation block.

\noindent\textbf{Force-aware sensor actions.}
Inspired by DexForce~\citep{chen2025dexforceextractingforceinformedactions},
we augment sensor-frame translation with a force-dependent
normal displacement to capture contact-loading changes
beyond observed motion.
For hand $h$ at frame $t$, let $(R_t^h,\mathbf{p}_t^h)$
denote the sensor pose and $f_t^h$ its estimated normal force.
We define
\begin{equation}
\begin{aligned}
    \Delta R_t^h
    &= (R_t^h)^\top R_{t+1}^h, \\
    \Delta\widetilde{\mathbf{p}}_t^h
    &= (R_t^h)^\top
       (\mathbf{p}_{t+1}^h-\mathbf{p}_t^h)
       + k_f\Delta f_t^h\mathbf{n}_s, \\
    \mathbf{a}_{t+1}^h
    &= [\Delta\widetilde{\mathbf{p}}_t^h;
        \operatorname{Rot6D}(\Delta R_t^h)]
       \in\mathbb{R}^{9},
\end{aligned}
\label{eq:sensor_action}
\end{equation}
where $\Delta f_t^h=f_{t+1}^h-f_t^h$,
$\mathbf{n}_s$ is the unit gel normal in sensor coordinates,
$k_f=0.5\,\mathrm{mm}/\mathrm{N}$, and $\operatorname{Rot6D}$ is the
6-D rotation encoding of \citet{zhou2019continuity}.

\noindent\textbf{View-aligned actions.}
Inspired by Action Images~\citep{zhen2026actionimagesendtoendpolicy},
we encode sensor position and orientation in the scene view
using three sensor-fixed points: the gel center and offsets
along the gel normal and an in-plane axis.
For each sensor $h$ and frame $t$, camera calibration maps
these points to normalized pixel coordinates and camera-space
depths, which we concatenate into
$\mathbf{u}_t^h\in\mathbb{R}^{9}$.
Depth retains geometric information lost during projection.
Appendix~\ref{app:action_representation} details the force
augmentation and reference-point projection.

\subsection{Multimodal Sequence Construction}
\label{sec:multimodal_transformer}

\noindent\textbf{Sequence construction.}
To jointly predict future observations from history and actions,
we pack clean history tokens, action tokens, and noisy future
target tokens from all streams into a single sequence.
We initialize the generator from pretrained
Cosmos3~\citep{nvidia2026cosmos3omnimodalworld} weights
to adapt its spatiotemporal representations to visuo-tactile prediction.
To preserve conditioning information, we corrupt only future
target latents and apply noise-level embeddings only to their tokens.

\noindent\textbf{Temporal alignment.}
To align observations and actions by physical time, we adopt
Cosmos3's multimodal rotary position embedding
(MRoPE)~\citep{nvidia2026cosmos3omnimodalworld}, which encodes
time, height, and width in separate rotary dimensions.
Each token is assigned a timestamp $\tau_i$, where observation
tokens use latent-block endpoints, sensor-action tokens use
transition endpoints, and view-aligned action tokens use
their corresponding frames.
We map these timestamps to a shared temporal axis using
Cosmos3's pretrained time scale, aligning streams in time
while preserving spatial structure within observations.

\noindent\textbf{Attention mask.}
To predict observation dynamics under a known action trajectory,
we allow all tokens to access the complete action window while
restricting attention to observation keys at the same or earlier
timestamps. We implement this through an additive attention mask:
\begin{equation}
    M_{ij} =
    \begin{cases}
        0, & j\in\mathcal{I}_{A}\ \text{or}\ \tau_j \leq \tau_i, \\
        -\infty, & \text{otherwise},
    \end{cases}
    \label{eq:causal_attention}
\end{equation}
where $i,j$ index query and key tokens within a sample,
$\tau_i$ denotes token time, and $\mathcal{I}_{A}$ contains
sensor-action and view-action token indices.
Applied in every layer, this mask preserves temporal ordering
in direct observation attention while allowing same-time tokens
to exchange information across streams for joint prediction.
To prevent ground-truth leakage, we additionally construct
each prediction prefix without later observation blocks or
clean copies of the current target in Sec.~\ref{sec:training_and_rollout}.

\subsection{Training and Inference}
\label{sec:training_and_rollout}

\noindent\textbf{Training.}
To learn sequential dynamics across observation streams,
we train ACTS with blockwise teacher forcing.
Each future block $q$ is predicted from clean observation
history, preceding ground-truth future blocks, and the full
action window, collectively denoted by $\mathcal{C}_q$.
Following Cosmos3~\citep{nvidia2026cosmos3omnimodalworld},
we use rectified flow matching~\citep{lipman2023flow,liu2023flow} with noisy targets
\begin{equation}
    \widetilde z_q^m
    = (1-\sigma)z_q^m+\sigma\epsilon_q^m,
    \qquad
    \epsilon_q^m\sim\mathcal{N}(0,I),
    \label{eq:flow_interpolation}
\end{equation}
where $m$ indexes observation streams and $\sigma$ is the
noise level.
The model jointly predicts velocities
$v_q^{m,*}=\epsilon_q^m-z_q^m$ by minimizing
\begin{equation}
    \mathcal{L}
    =
    \mathbb{E}\!\left[
        \frac{1}{Q}
        \sum_{q=1}^{Q}
        \sum_{m\in\mathcal{M}}
        \frac{
            \left\|
                v_\theta^m(\widetilde Z_q,\mathcal{C}_q,\sigma)
                -v_q^{m,*}
            \right\|_2^2
        }{D_z}
    \right],
    \label{eq:flow_loss}
\end{equation}
where $\widetilde Z_q$ collects the noisy target streams,
$Q$ is the number of future blocks, and $D_z$ is the number
of scalar elements per stream's latent block.
Only current target tokens are corrupted and supervised.

\noindent\textbf{Inference.}
To follow the sequential prediction process used during training,
we generate future latent blocks autoregressively within each
prediction window.
Each block is conditioned on the observed history, previously
generated blocks, and the complete action sequence.
For the first block, the observation context contains only
the recorded history.
We initialize the target latents of all streams with independent
Gaussian noise and jointly transform them into clean predictions
by integrating the learned velocity field, while keeping
the conditioning tokens fixed.
The completed block is then appended as clean latent context
for predicting the next block, replacing the ground-truth
context used during teacher forcing.
This process proceeds entirely in latent space without
intermediate decoding.
After all future blocks are generated, the corresponding decoders $\{\mathcal{D}_m\}_{m\in\mathcal{M}}$
reconstruct the predicted latents into synchronized scene,
wrist, and tactile observations.
Appendix~\ref{sec:implementation_details} provides
further training and inference details.

\section{Experiments}

\subsection{Baseline Models}
We compare ACTS with two action-conditioned visuo-tactile world
models, training both from scratch on React.
Among recent visuo-tactile world models, only
ContactWorld~\citep{zhang2026contactworld} and
Dream-Tac~\citep{lou2026dreamtac} release training code;
VT-WM~\citep{higuera2026visuotactileworldmodels},
OmniVTA~\citep{zheng2026omnivta}, ViTacWorld~\citep{huang2026vitacworld},
FeelWorld~\citep{ma2026feelworld}, and VT-WAM~\citep{tian2026vtwam} had
not released model code at the time of writing.
\textbf{ContactWorld} is a decoder-free JEPA-style~\citep{assran2023ijepa} model, which we
train with its official code, adapted to our observation and action
spaces.
\textbf{VT-WM-style} is our reimplementation of
VT-WM, preserving its
autoregressive latent predictor and action conditioning while
adapting its interfaces to our data.
We intended to compare with Dream-Tac but could not adapt it to
React: it generates actions jointly with future observations instead
of conditioning on given actions, and its released code expects a
robot camera layout and proprioception that React lacks.
Implementation details and deviations appear in
Appendices~\ref{sec:contactworld_baseline} and~\ref{app:vtwm}.

\subsection{Evaluation Metrics}
\label{sec:metrics}
We compare ACTS with baseline models on future-frame prediction
using held-out human recordings and zero-shot transfer to robot episodes.

\noindent\textbf{Video Generation Metrics.}
Given eight observed frames (one for ContactWorld, its native context)
and recorded actions, each model predicts the next eight frames (0.53\,s at 15\,fps).
We evaluate only predicted frames, separately for the scene view,
two wrist cameras, and two GelSight streams.
We report PSNR for pixel-level fidelity,
SSIM~\citep{1284395} for structural similarity, and
LPIPS~\citep{zhang2018perceptual} for perceptual distance.
Higher PSNR and SSIM and lower LPIPS indicate better results.
For both human and robot data, reference frames are decoded from
ground-truth latents using the frozen RGB and tactile VAEs.
All comparisons use identical test windows and seeds.
Appendix~\ref{app:evaluation_details} provides further details.

\noindent\textbf{Robot Experiments.}
We apply models trained on human data to held-out robot episodes
(Sec.~\ref{robot_data}) without fine-tuning.
Robot recordings are normalized to match the human data's tactile
color and illumination and wrist-camera tone.
Actions are constructed from OptiTrack sensor poses and
GelSight-estimated force following Sec.~\ref{sec:action_conditioning},
as for the human data, rather than from robot joint readings.
Per-step translations from forward kinematics differ from the
OptiTrack ones by a median of 0.33\,mm, about half the median
per-step motion, so OptiTrack keeps robot actions consistent with
the human training data
(Appendix~\ref{app:evaluation_details}).
We use the same eight-frame observation and prediction protocol
on windows in which contact is made or broken, and report the same
per-stream metrics.

\vspace{-2ex}
\section{Results}
\subsection{Implementation Details}
We train ACTS separately per task with the same architecture
and training recipe, initializing from
Cosmos3-Nano~\citep{nvidia2026cosmos3omnimodalworld}.
Observation VAEs and backbone weights remain frozen, while
45.3M parameters are trained through
LoRA~\citep{hu2021loralowrankadaptationlarge}
and modality-specific interfaces.
All ACTS models are trained for 20 epochs on four NVIDIA H100 GPUs,
with training, validation, and test splits separated by source episode.
Appendix~\ref{sec:implementation_details} provides
implementation and dataset details.

\subsection{Qualitative Examples}

\begin{figure}[t]
    \centering
    \vspace{-15 pt}
    \includegraphics[width=0.95\columnwidth]{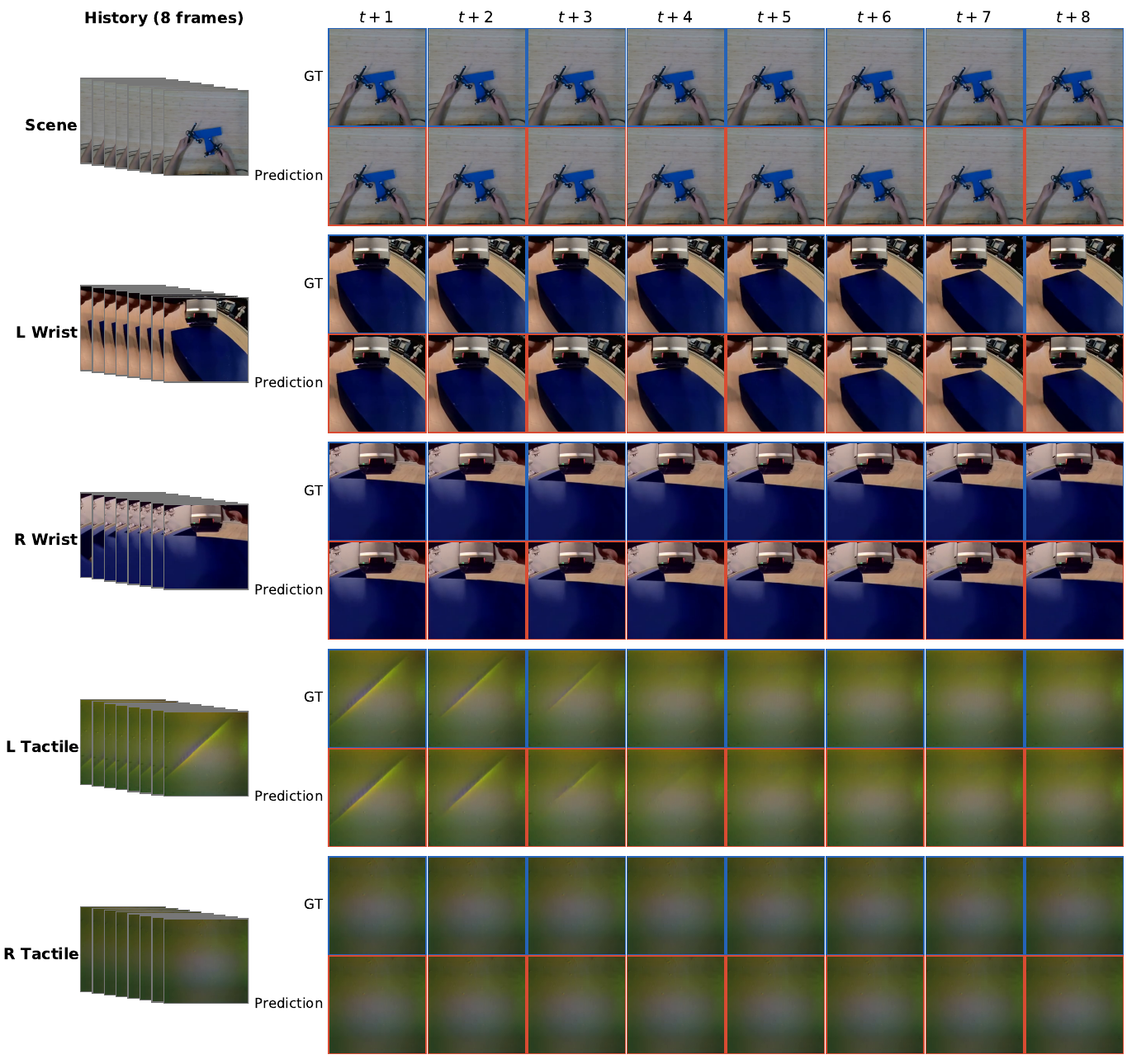}
    \caption{\textbf{Qualitative results} on a held-out PushT window. Given
    eight observed frames (left) and the recorded sensor actions, ACTS
    predicts the next eight frames of the scene view, both wrist views, and
    both GelSight streams. Ground truth in blue boxes, predictions in orange
    boxes. Videos and results on all tasks:
    \url{https://acts-anon.github.io/}.}
    \label{fig:qual}
    \vspace{-10 pt}
\end{figure}

Fig.~\ref{fig:qual} shows a held-out PushT window in which the operator lifts the left sensor off the block; more rollouts are on the project page. The recorded left imprint fades over $t{+}1$--$t{+}3$, and the predicted imprint fades on the same frames. The block stays in place, and the scene and wrist predictions remain consistent with the recording. Predicted imprints are slightly smoother than recorded ones, which we attribute to the tactile VAE.

\subsection{Quantitative Results}

Tab.~\ref{tab:quantative_results} compares ACTS with a persistence baseline that repeats the last observed frame. ACTS improves all camera streams by $1.0$--$4.6$\,dB PSNR, indicating that it captures scene motion and wrist-camera ego-motion beyond temporal persistence. On the scene view, ACTS also outperforms ContactWorld and VT-WM-style on all three objects in PSNR, SSIM, and LPIPS (Tab.~\ref{tab:baseline_comparison}).

Tactile scores require more careful interpretation. GelSight observations change only locally during contact, so a copied frame already achieves high PSNR and SSIM over the short prediction horizon. ACTS shows little consistent gain in full-frame tactile PSNR, although LPIPS improves on all tactile streams. These metrics capture appearance, not whether contact is made or broken at the right time (Sec.~\ref{sec:limitations}), so we also inspect rollouts. Appendix~\ref{app:force_ablation} ablates the force term in the sensor action (Eq.~\ref{eq:sensor_action}): it changes camera-stream PSNR by at most 0.3\,dB, never significantly, and tactile PSNR by at most 0.6\,dB, yet in the window of Fig.~\ref{fig:teaser}b it decides whether an imprint forms when the sensor reaches the block.

\begin{table*}[t]
\vspace{-10 pt}
\centering
\footnotesize
\setlength{\tabcolsep}{4pt}
\renewcommand{\arraystretch}{1.05}
\caption{Short-horizon prediction quality over 8 frames ($0.53$\,s).
\emph{Copy} repeats the last observed frame across the prediction window.
ACTS consistently improves the camera streams, while high temporal
persistence makes full-frame tactile metrics less discriminative.}
\label{tab:quantative_results}
\vspace{1ex}
\begin{tabular*}{\textwidth}{@{\extracolsep{\fill}} ll *{10}{r} @{}}
\toprule
& & \multicolumn{2}{c}{Video} & \multicolumn{2}{c}{Wrist (L)}
& \multicolumn{2}{c}{Wrist (R)} & \multicolumn{2}{c}{Tactile (L)}
& \multicolumn{2}{c}{Tactile (R)} \\
\cmidrule(lr){3-4}\cmidrule(lr){5-6}\cmidrule(lr){7-8}\cmidrule(lr){9-10}\cmidrule(lr){11-12}
Task & Metric & Copy & Ours & Copy & Ours & Copy & Ours & Copy & Ours & Copy & Ours \\
\midrule
\multirow{3}{*}{Rope}
& PSNR$\uparrow$     & 25.10 & \textbf{28.92} & 17.66 & \textbf{22.30} & 19.66 & \textbf{23.70} & 44.74 & \textbf{46.08} & 46.46 & \textbf{46.51} \\
& SSIM$\uparrow$     & 0.848 & \textbf{0.907} & 0.657 & \textbf{0.769} & 0.704 & \textbf{0.796} & 0.986 & \textbf{0.988} & 0.990 & \textbf{0.990} \\
& LPIPS$\downarrow$  & 0.115 & \textbf{0.045} & 0.261 & \textbf{0.131} & 0.234 & \textbf{0.129} & 0.033 & \textbf{0.021} & 0.027 & \textbf{0.020} \\
\midrule
\multirow{3}{*}{PushT}
& PSNR$\uparrow$     & 25.21 & \textbf{29.69} & 19.32 & \textbf{22.86} & 19.20 & \textbf{22.76} & 46.24 & \textbf{46.42} & \textbf{48.73} & 48.37 \\
& SSIM$\uparrow$     & 0.861 & \textbf{0.925} & 0.762 & \textbf{0.834} & 0.766 & \textbf{0.844} & 0.986 & \textbf{0.988} & 0.991 & \textbf{0.991} \\
& LPIPS$\downarrow$  & 0.103 & \textbf{0.038} & 0.176 & \textbf{0.097} & 0.194 & \textbf{0.111} & 0.027 & \textbf{0.022} & 0.016 & \textbf{0.015} \\
\midrule
\multirow{3}{*}{\shortstack[l]{Mother-\\board}}
& PSNR$\uparrow$     & 26.62 & \textbf{27.74} & 16.13 & \textbf{17.68} & 19.73 & \textbf{20.77} & \textbf{41.94} & 41.65 & \textbf{46.67} & 45.70 \\
& SSIM$\uparrow$     & 0.865 & \textbf{0.880} & 0.494 & \textbf{0.525} & 0.651 & \textbf{0.686} & \textbf{0.979} & 0.977 & \textbf{0.988} & 0.988 \\
& LPIPS$\downarrow$  & 0.065 & \textbf{0.046} & 0.307 & \textbf{0.258} & 0.188 & \textbf{0.162} & 0.043 & \textbf{0.041} & 0.031 & \textbf{0.029} \\
\bottomrule
\end{tabular*}
\vspace{-8 pt}
\end{table*}

\begin{table*}[t]
\centering
\footnotesize
\setlength{\tabcolsep}{2pt}
\renewcommand{\arraystretch}{1.08}
\caption{Scene-view prediction on held-out human data over 8 frames
($0.53$\,s). Both baselines are trained from scratch on React. Best
results per object are shown in bold.}
\label{tab:baseline_comparison}
\vspace{1ex}
\begin{tabular*}{\textwidth}{
    @{\extracolsep{\fill}}
    ll
    S[table-format=2.2] S[table-format=1.4] S[table-format=1.4]
    @{}
}
\toprule
Task & Model
& {PSNR$\uparrow$} & {SSIM$\uparrow$} & {LPIPS$\downarrow$} \\
\midrule

\multirow{3}{*}{Rope}
& ContactWorld~\citeyearpar{zhang2026contactworld}
& 26.03 & 0.846 & 0.164 \\
& VT-WM-style~\citeyearpar{higuera2026visuotactileworldmodels}
& 28.05 & 0.8878 & 0.1118 \\
& Ours
& {\textbf{28.92}} & {\textbf{0.907}} & {\textbf{0.045}} \\
\midrule

\multirow{3}{*}{PushT}
& ContactWorld~\citeyearpar{zhang2026contactworld}
& 26.35 & 0.858 & 0.139 \\
& VT-WM-style~\citeyearpar{higuera2026visuotactileworldmodels}
& 27.69 & 0.8916 & 0.1047 \\
& Ours
& {\textbf{29.69}} & {\textbf{0.925}} & {\textbf{0.038}} \\
\midrule

\multirow{3}{*}{Motherboard}
& ContactWorld~\citeyearpar{zhang2026contactworld}
& 25.85 & 0.835 & 0.120 \\
& VT-WM-style~\citeyearpar{higuera2026visuotactileworldmodels}
& 27.01 & 0.8692 & 0.0837 \\
& Ours
& {\textbf{27.74}} & {\textbf{0.880}} & {\textbf{0.046}} \\

\bottomrule
\end{tabular*}
\vspace{-10 pt}
\end{table*}

\subsection{Zero-Shot Transfer to Robot Teleoperation}
\label{sec:robot_transfer}

Tab.~\ref{tab:robot_results} applies the same models, without fine-tuning, to held-out robot teleoperation, scored as in Tab.~\ref{tab:quantative_results} against VAE-reconstructed ground truth. Robot motion is slower than human manipulation, so repeating the last frame is a stronger baseline than on human data. Transfer varies by object. On PushT, ACTS improves all three camera streams in PSNR and SSIM, by up to $2.1$\,dB PSNR on the scene view. On Rope, the results are mixed: scene PSNR rises by $0.7$\,dB while right-wrist PSNR drops by $0.4$\,dB. On Motherboard, ACTS falls below frame copying on every camera stream. Likely causes are its geometry, whose connectors, slots, and heat sinks put many fine edges in view, and the robot wrist cameras, which record it darker than the human data. Across objects, the wrist views degrade most: predictions are blurrier than on human data and sometimes contain object structure absent from the recording. On the tactile streams, frame copying scores higher PSNR on five of the six streams, by up to $2.1$\,dB, while ACTS scores lower LPIPS on five of the six.

Fig.~\ref{fig:robot_qual} (Appendix~\ref{app:robot_qual}) shows a held-out robot PushT window in which the sensor breaks contact with the block. The predicted left GelSight imprint fades within one frame of the recorded one, and the scene prediction keeps the block in place. The left wrist view shows the failure described above: the recorded view barely changes, while the predicted block face grows and shifts from $t{+}4$ onward. Across the robot rollouts we inspected, including those on the project page, predicted imprints appear and disappear as contact is made and broken in the recording, even where the wrist predictions degrade.

\begin{table*}[t]
\vspace{-15 pt}
\centering
\footnotesize
\setlength{\tabcolsep}{4pt}
\renewcommand{\arraystretch}{1.05}
\caption{Zero-shot short-horizon prediction on robot teleoperation (8 frames, $0.53$\,s), scored
against VAE-reconstructed ground truth. \emph{Copy} repeats the last observed frame; slow robot
motion makes it a strong baseline.}
\label{tab:robot_results}
\vspace{1ex}
\begin{tabular*}{\textwidth}{@{\extracolsep{\fill}} ll *{10}{r} @{}}
\toprule
& & \multicolumn{2}{c}{Video} & \multicolumn{2}{c}{Wrist (L)}
& \multicolumn{2}{c}{Wrist (R)} & \multicolumn{2}{c}{Tactile (L)}
& \multicolumn{2}{c}{Tactile (R)} \\
\cmidrule(lr){3-4}\cmidrule(lr){5-6}\cmidrule(lr){7-8}\cmidrule(lr){9-10}\cmidrule(lr){11-12}
Task & Metric & Copy & Ours & Copy & Ours & Copy & Ours & Copy & Ours & Copy & Ours \\
\midrule
\multirow{3}{*}{Rope}
& PSNR$\uparrow$     & 31.57 & \textbf{32.31} & 27.29 & \textbf{27.88} & \textbf{27.70} & 27.28 & \textbf{42.74} & 42.56 & \textbf{44.73} & 44.65 \\
& SSIM$\uparrow$     & \textbf{0.944} & 0.942 & 0.827 & \textbf{0.845} & 0.838 & \textbf{0.856} & 0.980 & \textbf{0.980} & 0.986 & \textbf{0.986} \\
& LPIPS$\downarrow$  & \textbf{0.021} & 0.022 & 0.055 & \textbf{0.055} & \textbf{0.054} & 0.057 & 0.043 & \textbf{0.029} & 0.040 & \textbf{0.025} \\
\midrule
\multirow{3}{*}{PushT}
& PSNR$\uparrow$     & 27.75 & \textbf{29.85} & 25.84 & \textbf{26.71} & 23.96 & \textbf{25.62} & \textbf{45.66} & 44.15 & \textbf{45.49} & 44.25 \\
& SSIM$\uparrow$     & 0.915 & \textbf{0.924} & 0.850 & \textbf{0.863} & 0.827 & \textbf{0.853} & 0.980 & \textbf{0.982} & 0.984 & \textbf{0.984} \\
& LPIPS$\downarrow$  & 0.035 & \textbf{0.030} & \textbf{0.044} & 0.048 & 0.064 & \textbf{0.053} & 0.031 & \textbf{0.024} & 0.040 & \textbf{0.032} \\
\midrule
\multirow{3}{*}{\shortstack[l]{Mother-\\board}}
& PSNR$\uparrow$     & \textbf{32.46} & 30.63 & \textbf{26.72} & 25.71 & \textbf{25.88} & 24.92 & \textbf{46.79} & 44.65 & 44.39 & \textbf{44.60} \\
& SSIM$\uparrow$     & \textbf{0.941} & 0.922 & \textbf{0.819} & 0.753 & \textbf{0.817} & 0.811 & \textbf{0.990} & 0.987 & 0.986 & \textbf{0.987} \\
& LPIPS$\downarrow$  & \textbf{0.015} & 0.024 & \textbf{0.050} & 0.073 & \textbf{0.052} & 0.059 & \textbf{0.016} & 0.023 & 0.035 & \textbf{0.025} \\
\bottomrule
\vspace{-15 pt} 
\end{tabular*}
\end{table*}

\section{Limitations}\label{sec:limitations}

\noindent\textbf{Force estimation.}
The estimator predicts only normal force, which governs making and breaking contact but not the shear that governs sliding. Its error of 0.318\,N (6.4\%) is measured on CNC calibration data (Appendix~\ref{sec:normal_force_estimation}); the human recordings have no force ground truth, so its error during manipulation is unmeasured.

\noindent\textbf{Evaluation metrics.}
All streams are scored against VAE-reconstructed ground truth, so errors shared by the model and the VAE go undetected. On GelSight images, full-frame metrics are dominated by static gel and do not show whether contact is made or broken at the right time, which we assess only qualitatively.

\noindent\textbf{Motion capture.}
Sensor poses come from OptiTrack for both human collection and robot evaluation, which confines React to one instrumented lab. Forward kinematics is not a substitute: its per-step translations differ from OptiTrack by a median of 0.33\,mm, about half the median per-step motion and equal to the force term for a 0.66\,N change (Appendix~\ref{app:evaluation_details}). Robot data collection and deployment therefore also need external tracking.

\section{Conclusion}

We presented ACTS, a visuo-tactile world model trained entirely on human physical interaction, with actions defined at the tactile sensor as motion augmented with force change. It predicts the scene view better than two visuo-tactile baselines on held-out human data, and on zero-shot robot teleoperation its tactile imprints track contact transitions while wrist views degrade. Improving visual transfer and using ACTS for robot planning are future work.

\subsection*{Ethics statement}
React records five human operators who are members of the research team and gave informed consent to the recording and release of the data. The third-person cameras capture the operators' hands and forearms, and no frame captures an operator's face. The dataset contains no personal information beyond these recordings. The objects are inanimate, and the robot experiments involve no human subjects. We see no application of this work that raises concerns beyond those common to robot manipulation research.

\subsection*{Reproducibility statement}

Appendix~\ref{app:dataset} describes data collection, stream alignment,
curation, and normal-force estimation, with dataset statistics in
Tab.~\ref{tab:dataset_statistics}. Appendix~\ref{sec:implementation_details}
gives the model architecture, training hyperparameters, data splits, and
compute, and Appendix~\ref{sec:tactile_encoder} the tactile VAE adaptation.
Baseline configurations and every deviation from the published models are
listed in Appendices~\ref{sec:contactworld_baseline} and~\ref{app:vtwm}.
Evaluation protocol and metrics are defined in Sec.~\ref{sec:metrics}, and
all comparisons use identical test windows and seeds. We will release the
React dataset, the trained models, and the training and evaluation code upon
acceptance.

\subsection*{Use of generative AI}

In this work, we used generative AI tools (Claude, Anthropic) to provide
feedback on experimental design, to help interpret results, and to help build
the automatic pipeline that cleans and reformats raw recordings into the
dataset.
We did not use generative AI tools to generate synthetic data, develop
theoretical models or conceptual frameworks, propose or refine hypotheses,
implement methods, or translate; formulating or proving mathematical claims
and qualitative or thematic data analysis are not applicable to this work.
Additionally, we used generative AI tools to create or edit software code,
draft and edit parts of the paper, and identify and format references.
We have reviewed all AI-assisted work. Experimental changes suggested by the AI
tool, such as additional baselines and contact-timing metrics, were treated as
proposals, and the authors decided which to pursue. The curation pipeline applies fixed rules, removing
intervals with camera freezes or invalid pose tracking, and keeps the source-frame
index of every aligned sample so that its output can be audited against the raw
recordings (Appendix~\ref{app:dataset}). Every description of a qualitative result was checked
against the corresponding figure by the authors, and every explanation of a
quantitative result reflects the authors' own analysis. Bibliography entries
added with AI assistance were retrieved from their DOIs and checked against
CrossRef, and no reference was written from the model's memory.
We take responsibility for the final content of this work, including text,
claims or artifacts produced with the aid of generative AI.

\bibliography{iclr2027_conference}

\begin{thebibliography}{52}
\providecommand{\natexlab}[1]{#1}
\providecommand{\url}[1]{\texttt{#1}}
\expandafter\ifx\csname urlstyle\endcsname\relax
  \providecommand{\doi}[1]{doi: #1}\else
  \providecommand{\doi}{doi: \begingroup \urlstyle{rm}\Url}\fi

\bibitem[Assran et~al.(2023)Assran, Duval, Misra, Bojanowski, Vincent, Rabbat,
  LeCun, and Ballas]{assran2023ijepa}
Mahmoud Assran, Quentin Duval, Ishan Misra, Piotr Bojanowski, Pascal Vincent,
  Michael Rabbat, Yann LeCun, and Nicolas Ballas.
\newblock Self-supervised learning from images with a joint-embedding
  predictive architecture.
\newblock In \emph{Proceedings of the IEEE/CVF Conference on Computer Vision
  and Pattern Recognition (CVPR)}, 2023.

\bibitem[Chen et~al.(2025)Chen, Yu, Choi, Cutkosky, and
  Bohg]{chen2025dexforceextractingforceinformedactions}
Claire Chen, Zhongchun Yu, Hojung Choi, Mark Cutkosky, and Jeannette Bohg.
\newblock Dexforce: Extracting force-informed actions from kinesthetic
  demonstrations for dexterous manipulation, 2025.
\newblock URL \url{https://arxiv.org/abs/2501.10356}.

\bibitem[Cheng et~al.(2026)Cheng, Chen, Chen, Zhang, Zhang, Ling, Hamad, Bing,
  Wu, Sharma, and Knoll]{cheng2026tacumi}
Tailai Cheng, Kejia Chen, Lingyun Chen, Liding Zhang, Yue Zhang, Yao Ling,
  Mahdi Hamad, Zhenshan Bing, Fan Wu, Karan Sharma, and Alois Knoll.
\newblock {TacUMI}: A multi-modal universal manipulation interface for
  contact-rich tasks.
\newblock In \emph{2026 IEEE International Conference on Robotics and
  Automation (ICRA)}. IEEE, 2026.

\bibitem[Chi et~al.(2024)Chi, Xu, Pan, Cousineau, Burchfiel, Feng, Tedrake, and
  Song]{chi2024universal}
Cheng Chi, Zhenjia Xu, Chuer Pan, Eric Cousineau, Benjamin Burchfiel, Siyuan
  Feng, Russ Tedrake, and Shuran Song.
\newblock Universal manipulation interface: In-the-wild robot teaching without
  in-the-wild robots.
\newblock In \emph{Proceedings of Robotics: Science and Systems (RSS)}, 2024.

\bibitem[Dasari et~al.(2020)Dasari, Ebert, Tian, Nair, Bucher, Schmeckpeper,
  Singh, Levine, and Finn]{dasari2020robonet}
Sudeep Dasari, Frederik Ebert, Stephen Tian, Suraj Nair, Bernadette Bucher,
  Karl Schmeckpeper, Siddharth Singh, Sergey Levine, and Chelsea Finn.
\newblock {RoboNet}: Large-scale multi-robot learning.
\newblock In \emph{Proceedings of the Conference on Robot Learning}, volume 100
  of \emph{Proceedings of Machine Learning Research}, pp.\  885--897. PMLR,
  2020.

\bibitem[Dou et~al.(2024)Dou, Yang, Liu, Loquercio, and Owens]{dou2024tarf}
Yiming Dou, Fengyu Yang, Yi~Liu, Antonio Loquercio, and Andrew Owens.
\newblock Tactile-augmented radiance fields.
\newblock In \emph{Proceedings of the IEEE/CVF Conference on Computer Vision
  and Pattern Recognition}, pp.\  26529--26539, 2024.

\bibitem[Feng et~al.(2025)Feng, Hu, Xia, Gao, Shen, Sun, Fang, and
  Hu]{feng2025learning}
Ruoxuan Feng, Jiangyu Hu, Wenke Xia, Tianci Gao, Ao~Shen, Yuhao Sun, Bin Fang,
  and Di~Hu.
\newblock {AnyTouch}: Learning unified static-dynamic representation across
  multiple visuo-tactile sensors.
\newblock In \emph{The Thirteenth International Conference on Learning
  Representations}, 2025.

\bibitem[Finn \& Levine(2017)Finn and Levine]{finn2017deep}
Chelsea Finn and Sergey Levine.
\newblock Deep visual foresight for planning robot motion.
\newblock In \emph{2017 IEEE International Conference on Robotics and
  Automation (ICRA)}, pp.\  2786--2793. IEEE, 2017.
\newblock \doi{10.1109/ICRA.2017.7989324}.

\bibitem[Gao et~al.(2022)Gao, Si, Chang, Clarke, Bohg, Fei-Fei, Yuan, and
  Wu]{gao2022objectfolder}
Ruohan Gao, Zilin Si, Yen-Yu Chang, Samuel Clarke, Jeannette Bohg, Li~Fei-Fei,
  Wenzhen Yuan, and Jiajun Wu.
\newblock {ObjectFolder 2.0}: A multisensory object dataset for sim2real
  transfer.
\newblock In \emph{Proceedings of the IEEE/CVF Conference on Computer Vision
  and Pattern Recognition}, pp.\  10598--10608, 2022.

\bibitem[Guo et~al.(2026)Guo, Shi, Chen, and Finn]{guo2025ctrl}
Yanjiang Guo, Lucy~Xiaoyang Shi, Jianyu Chen, and Chelsea Finn.
\newblock {Ctrl-World}: A controllable generative world model for robot
  manipulation.
\newblock In \emph{International Conference on Learning Representations
  (ICLR)}, 2026.

\bibitem[Harley et~al.(2025)Harley, You, Sun, Zheng, Raghuraman, Gu, Liang,
  Chu, Dave, Tokmakov, You, Ambrus, Fragkiadaki, and
  Guibas]{harley2025alltrackerefficientdensepoint}
Adam~W. Harley, Yang You, Xinglong Sun, Yang Zheng, Nikhil Raghuraman, Yunqi
  Gu, Sheldon Liang, Wen-Hsuan Chu, Achal Dave, Pavel Tokmakov, Suya You, Rares
  Ambrus, Katerina Fragkiadaki, and Leonidas~J. Guibas.
\newblock {AllTracker}: Efficient dense point tracking at high resolution.
\newblock In \emph{Proceedings of the IEEE/CVF International Conference on
  Computer Vision (ICCV)}, 2025.

\bibitem[Higuera et~al.(2025)Higuera, Sharma, Fan, Bodduluri, Boots, Kaess,
  Lambeta, Wu, Liu, Hogan, and Mukadam]{higuera2025tactile}
Carolina Higuera, Akash Sharma, Taosha Fan, Chaithanya~Krishna Bodduluri, Byron
  Boots, Michael Kaess, Mike Lambeta, Tingfan Wu, Zixi Liu, Francois~Robert
  Hogan, and Mustafa Mukadam.
\newblock Tactile beyond pixels: Multisensory touch representations for robot
  manipulation.
\newblock In \emph{Proceedings of The 9th Conference on Robot Learning}, volume
  305 of \emph{Proceedings of Machine Learning Research}, pp.\  105--123, 2025.

\bibitem[Higuera et~al.(2026)Higuera, Arnaud, Boots, Mukadam, Hogan, and
  Meier]{higuera2026visuotactileworldmodels}
Carolina Higuera, Sergio Arnaud, Byron Boots, Mustafa Mukadam, Francois~Robert
  Hogan, and Franziska Meier.
\newblock Visuo-tactile world models, 2026.
\newblock URL \url{https://arxiv.org/abs/2602.06001}.

\bibitem[Hou et~al.(2026)Hou, Li, Jia, An, Guo, Leng, Geng, Ze, Harada, Torr,
  et~al.]{hou2026worldmodelsurvey}
Bohan Hou, Gen Li, Jindou Jia, Tuo An, Xinying Guo, Sicong Leng, Haoran Geng,
  Yanjie Ze, Tatsuya Harada, Philip Torr, et~al.
\newblock World model for robot learning: A comprehensive survey.
\newblock \emph{arXiv preprint arXiv:2605.00080}, 2026.

\bibitem[Hu et~al.(2022)Hu, Shen, Wallis, Allen-Zhu, Li, Wang, Wang, and
  Chen]{hu2021loralowrankadaptationlarge}
Edward~J. Hu, Yelong Shen, Phillip Wallis, Zeyuan Allen-Zhu, Yuanzhi Li, Shean
  Wang, Lu~Wang, and Weizhu Chen.
\newblock {LoRA}: Low-rank adaptation of large language models.
\newblock In \emph{International Conference on Learning Representations
  (ICLR)}, 2022.

\bibitem[Huang et~al.(2025)Huang, Kaess, and Yuan]{huang2025normalflow}
Hung-Jui Huang, Michael Kaess, and Wenzhen Yuan.
\newblock {NormalFlow}: Fast, robust, and accurate contact-based object {6DoF}
  pose tracking with vision-based tactile sensors.
\newblock \emph{IEEE Robotics and Automation Letters}, 10\penalty0
  (1):\penalty0 452--459, 2025.
\newblock \doi{10.1109/LRA.2024.3505815}.

\bibitem[Huang et~al.(2026{\natexlab{a}})Huang, Mirzaee, Kaess, and
  Yuan]{huang2025gelslam}
Hung-Jui Huang, Mohammad~Amin Mirzaee, Michael Kaess, and Wenzhen Yuan.
\newblock {GelSLAM}: A real-time, high-fidelity, and robust 3d tactile slam
  system.
\newblock \emph{IEEE Transactions on Robotics}, 2026{\natexlab{a}}.
\newblock \doi{10.1109/TRO.2026.3734131}.

\bibitem[Huang et~al.(2026{\natexlab{b}})Huang, Sang, Ni, Lu, Wu, Lv, Guo, Shi,
  and Wang]{huang2026vitacworld}
Yunao Huang, Shiyu Sang, Suting Ni, Haotao Lu, Shijie Wu, Rui Lv, Ziyang Guo,
  Ye~Shi, and Jingya Wang.
\newblock {ViTacWorld}: Scaling visuo-tactile world models for contact-rich
  robot manipulation.
\newblock \emph{arXiv preprint arXiv:2607.22530}, 2026{\natexlab{b}}.
\newblock URL \url{https://arxiv.org/abs/2607.22530}.

\bibitem[Johnson \& Adelson(2009)Johnson and Adelson]{johnson2009retrographic}
Micah~K. Johnson and Edward~H. Adelson.
\newblock Retrographic sensing for the measurement of surface texture and
  shape.
\newblock In \emph{IEEE Conference on Computer Vision and Pattern Recognition
  (CVPR)}, pp.\  1070--1077. IEEE, 2009.
\newblock \doi{10.1109/CVPR.2009.5206534}.

\bibitem[Kota et~al.(2026)Kota, Shah, Colgate, and Reardon]{kota2026threedcal}
Rohan Kota, Kaival Shah, J.~Edward Colgate, and Gregory Reardon.
\newblock {3D Cal}: An open-source software library for depth reconstruction on
  vision-based tactile sensors.
\newblock \emph{IEEE Robotics and Automation Letters}, 11\penalty0
  (5):\penalty0 6248--6255, 2026.
\newblock \doi{10.1109/LRA.2026.3673994}.

\bibitem[Krizhevsky et~al.(2012)Krizhevsky, Sutskever, and
  Hinton]{Krizhevsky2012ImageNet}
Alex Krizhevsky, Ilya Sutskever, and Geoffrey~E. Hinton.
\newblock Imagenet classification with deep convolutional neural networks.
\newblock In F.~Pereira, C.~J.~C. Burges, L.~Bottou, and K.~Q. Weinberger
  (eds.), \emph{Advances in Neural Information Processing Systems}, volume~25,
  pp.\  1097--1105. Curran Associates, Inc., 2012.

\bibitem[Lambeta et~al.(2024)Lambeta, Wu, Sengul, Most, Black, Sawyer, Mercado,
  Qi, Sohn, Taylor, Tydingco, Kammerer, Stroud, Khatha, Jenkins, Most, Stein,
  Chavira, Craven-Bartle, Sanchez, Ding, Malik, and
  Calandra]{lambeta2024digitizing}
Mike Lambeta, Tingfan Wu, Ali Sengul, Victoria~Rose Most, Nolan Black, Kevin
  Sawyer, Romeo Mercado, Haozhi Qi, Alexander Sohn, Byron Taylor, Norb
  Tydingco, Gregg Kammerer, Dave Stroud, Jake Khatha, Kurt Jenkins, Kyle Most,
  Neal Stein, Ricardo Chavira, Thomas Craven-Bartle, Eric Sanchez, Yitian Ding,
  Jitendra Malik, and Roberto Calandra.
\newblock Digitizing touch with an artificial multimodal fingertip.
\newblock \emph{arXiv preprint arXiv:2411.02479}, 2024.

\bibitem[Lipman et~al.(2023)Lipman, Chen, Ben-Hamu, Nickel, and
  Le]{lipman2023flow}
Yaron Lipman, Ricky T.~Q. Chen, Heli Ben-Hamu, Maximilian Nickel, and Matt Le.
\newblock Flow matching for generative modeling.
\newblock In \emph{International Conference on Learning Representations
  (ICLR)}, 2023.

\bibitem[Liu et~al.(2023)Liu, Gong, and Liu]{liu2023flow}
Xingchao Liu, Chengyue Gong, and Qiang Liu.
\newblock Flow straight and fast: Learning to generate and transfer data with
  rectified flow.
\newblock In \emph{International Conference on Learning Representations
  (ICLR)}, 2023.

\bibitem[Loshchilov \& Hutter(2019)Loshchilov and
  Hutter]{loshchilov2019decoupledweightdecayregularization}
Ilya Loshchilov and Frank Hutter.
\newblock Decoupled weight decay regularization.
\newblock In \emph{International Conference on Learning Representations
  (ICLR)}, 2019.

\bibitem[Lou et~al.(2026)Lou, Ye, Fu, Cen, Chi, Lyu, Jia, Han, Lu, and
  Zhang]{lou2026dreamtac}
Yunfan Lou, Yifan Ye, Yankai Fu, Jun Cen, Xiaowei Chi, Yaoxu Lyu, Peidong Jia,
  Sirui Han, Zhihe Lu, and Shanghang Zhang.
\newblock {Dream-Tac}: A unified tactile world action model for contact-rich
  robot manipulation.
\newblock \emph{arXiv preprint arXiv:2606.08737}, 2026.

\bibitem[Ma et~al.(2026)Ma, Zhang, Xue, Cai, Yao, Cui, and
  Wang]{ma2026feelworld}
Wenxuan Ma, Chaofan Zhang, Chao Xue, Yinghao Cai, Guocai Yao, Shaowei Cui, and
  Shuo Wang.
\newblock {FeelWorld}: Visuo-tactile world model for hierarchical contact
  prediction and planning, 2026.
\newblock URL \url{https://arxiv.org/abs/2607.24267}.

\bibitem[{NVIDIA}(2026)]{nvidia2026cosmos3omnimodalworld}
{NVIDIA}.
\newblock Cosmos 3: Omnimodal world models for physical ai, 2026.
\newblock URL \url{https://arxiv.org/abs/2606.02800}.

\bibitem[Schneider et~al.(2025)Schneider, Duret, de~Farias, Calandra, Chen, and
  Peters]{schneider2025tactilemnist}
Tim Schneider, Guillaume Duret, Cristiana de~Farias, Roberto Calandra, Liming
  Chen, and Jan Peters.
\newblock {Tactile MNIST}: Benchmarking active tactile perception, 2025.
\newblock URL \url{https://arxiv.org/abs/2506.06361}.

\bibitem[Shahidzadeh et~al.(2025)Shahidzadeh, Caddeo, Alapati, Natale,
  Ferm{\"u}ller, and Aloimonos]{shahidzadeh2024feelanyforce}
Amir-Hossein Shahidzadeh, Gabriele~M. Caddeo, Koushik Alapati, Lorenzo Natale,
  Cornelia Ferm{\"u}ller, and Yiannis Aloimonos.
\newblock {FeelAnyForce}: Estimating contact force feedback from tactile
  sensation for vision-based tactile sensors.
\newblock In \emph{IEEE International Conference on Robotics and Automation
  (ICRA)}, pp.\  251--257, 2025.
\newblock \doi{10.1109/ICRA55743.2025.11127723}.

\bibitem[Shan et~al.(2026)Shan, Zhou, Wang, Chen, Chen, Zhou, Ma, Hu, Li, Hu,
  Li, Chen, Zhu, Li, Ze, Geng, Dou, Bi, Liu, Zhou, Li, Liang, Harada,
  Katzschmann, Soh, Li, Johns, Kragic, Peters, Matusik, Tomizuka, Malik, and
  Yang]{shan2026learningphysicalinteractionsurvey}
Shilin Shan, Chuhao Zhou, Ruize Wang, Xinyan Chen, Xiangyu Chen, Xinyu Zhou,
  Boyu Ma, Iris~Yuxuan Hu, Jingliang Li, Celeste~Yuxuan Hu, Geng Li, Guohao
  Chen, Tianrui Zhu, Zhe Li, Yanjie Ze, Haoran Geng, Zhiyang Dou, Jianxin Bi,
  Yuejiang Liu, Jianshu Zhou, Jiachen Li, Paul Liang, Tatsuya Harada, Robert
  Katzschmann, Harold Soh, Na~Li, Edward Johns, Danica Kragic, Jan Peters,
  Wojciech Matusik, Masayoshi Tomizuka, Jitendra Malik, and Jianfei Yang.
\newblock Learning physical interaction: A survey of tactile- and force-aware
  robot learning, 2026.

\bibitem[Suresh et~al.(2024)Suresh, Qi, Wu, Fan, Pineda, Lambeta, Malik,
  Kalakrishnan, Calandra, Kaess, Ortiz, and Mukadam]{suresh2024neuralfeels}
Sudharshan Suresh, Haozhi Qi, Tingfan Wu, Taosha Fan, Luis Pineda, Mike
  Lambeta, Jitendra Malik, Mrinal Kalakrishnan, Roberto Calandra, Michael
  Kaess, Joseph Ortiz, and Mustafa Mukadam.
\newblock {NeuralFeels} with neural fields: Visuotactile perception for in-hand
  manipulation.
\newblock \emph{Science Robotics}, 9\penalty0 (96):\penalty0 eadl0628, 2024.
\newblock \doi{10.1126/scirobotics.adl0628}.

\bibitem[Tang et~al.(2026)Tang, Ji, Xing, Hao, Xu, Shao, Lu, Yu, Pang, and
  Zhang]{tang2026towards}
Tutian Tang, Xingyu Ji, Wanli Xing, Ce~Hao, Wenqiang Xu, Lin Shao, Cewu Lu,
  Qiaojun Yu, Jiangmiao Pang, and Kaifeng Zhang.
\newblock Towards human-like manipulation through rl-augmented teleoperation
  and mixture-of-dexterous-experts vla.
\newblock \emph{arXiv preprint arXiv:2603.08122}, 2026.

\bibitem[Tian et~al.(2026)Tian, Zheng, Zheng, Gu, Zang, Qin, Li, Li, Ding, and
  Zhao]{tian2026vtwam}
Shuai Tian, Yupeng Zheng, Yuhang Zheng, Songen Gu, Yujie Zang, Yuxing Qin,
  Weize Li, Haoran Li, Wenchao Ding, and Dongbin Zhao.
\newblock {VT-WAM}: Visual-tactile world action model for contact-rich
  manipulation, 2026.
\newblock URL \url{https://arxiv.org/abs/2607.02503}.

\bibitem[Wan et~al.(2025)Wan, Wang, Ai, Wen, Mao, Xie, Chen, Yu, Zhao, Yang,
  Zeng, Wang, Zhang, Zhou, Wang, Chen, Zhu, Zhao, Yan, Huang, Feng, Zhang, Li,
  Wu, Chu, Feng, Zhang, Sun, Fang, Wang, Gui, Weng, Shen, Lin, Wang, Wang,
  Zhou, Wang, Shen, Yu, Shi, Huang, Xu, Kou, Lv, Li, Liu, Wang, Zhang, Huang,
  Li, Wu, Liu, Pan, Zheng, Hong, Shi, Feng, Jiang, Han, Wu, and Liu]{wan2025}
Team Wan, Ang Wang, Baole Ai, Bin Wen, Chaojie Mao, Chen-Wei Xie, Di~Chen,
  Feiwu Yu, Haiming Zhao, Jianxiao Yang, Jianyuan Zeng, Jiayu Wang, Jingfeng
  Zhang, Jingren Zhou, Jinkai Wang, Jixuan Chen, Kai Zhu, Kang Zhao, Keyu Yan,
  Lianghua Huang, Mengyang Feng, Ningyi Zhang, Pandeng Li, Pingyu Wu, Ruihang
  Chu, Ruili Feng, Shiwei Zhang, Siyang Sun, Tao Fang, Tianxing Wang, Tianyi
  Gui, Tingyu Weng, Tong Shen, Wei Lin, Wei Wang, Wei Wang, Wenmeng Zhou, Wente
  Wang, Wenting Shen, Wenyuan Yu, Xianzhong Shi, Xiaoming Huang, Xin Xu, Yan
  Kou, Yangyu Lv, Yifei Li, Yijing Liu, Yiming Wang, Yingya Zhang, Yitong
  Huang, Yong Li, You Wu, Yu~Liu, Yulin Pan, Yun Zheng, Yuntao Hong, Yupeng
  Shi, Yutong Feng, Zeyinzi Jiang, Zhen Han, Zhi-Fan Wu, and Ziyu Liu.
\newblock Wan: Open and advanced large-scale video generative models.
\newblock \emph{arXiv preprint arXiv:2503.20314}, 2025.

\bibitem[Wang et~al.(2026)Wang, Wang, Pei, Zhang, Liang, Hu, Li, Wu, Han,
  Zhang, Qi, Wu, Zhang, Zheng, Pan, Navarro-Alarcon, Liu, and
  Zhou]{wang2026worldmodelsroboticmanipulation}
Fangyuan Wang, Ziyuan Wang, Guorui Pei, Mengshi Zhang, Canxi Liang, Jun Hu,
  Zhongxuan Li, Jinsong Wu, Ning Han, Zeqing Zhang, Jiaming Qi, Hongmin Wu,
  Shiyao Zhang, Pai Zheng, Jia Pan, David Navarro-Alarcon, Sichao Liu, and Peng
  Zhou.
\newblock World models for robotic manipulation: A survey.
\newblock \emph{SmartBot}, pp.\  e70053, 2026.
\newblock \doi{10.1002/smb2.70053}.

\bibitem[Wang et~al.(2004)Wang, Bovik, Sheikh, and Simoncelli]{1284395}
Zhou Wang, A.C. Bovik, H.R. Sheikh, and E.P. Simoncelli.
\newblock Image quality assessment: from error visibility to structural
  similarity.
\newblock \emph{IEEE Transactions on Image Processing}, 13\penalty0
  (4):\penalty0 600--612, 2004.
\newblock \doi{10.1109/TIP.2003.819861}.

\bibitem[Wu et~al.(2026)Wu, Yu, Ren, Chen, Jiang, Huang, Gu, and
  Li]{wu2026freetacman}
Longyan Wu, Checheng Yu, Jieji Ren, Li~Chen, Yufei Jiang, Ran Huang, Guoying
  Gu, and Hongyang Li.
\newblock {FreeTacMan}: Robot-free visuo-tactile data collection system for
  contact-rich manipulation.
\newblock In \emph{IEEE International Conference on Robotics and Automation
  (ICRA)}, pp.\  5367--5374, 2026.
\newblock \doi{10.1109/ICRA57385.2026.11696142}.

\bibitem[Wu et~al.(2024)Wu, Shentu, Yi, Lin, and Abbeel]{wu2024gello}
Philipp Wu, Yide Shentu, Zhongke Yi, Xingyu Lin, and Pieter Abbeel.
\newblock {GELLO}: A general, low-cost, and intuitive teleoperation framework
  for robot manipulators.
\newblock In \emph{2024 IEEE/RSJ International Conference on Intelligent Robots
  and Systems (IROS)}, pp.\  12156--12163. IEEE, 2024.

\bibitem[Xu et~al.(2025)Xu, Wei, An, Zhang, and Li]{xu2025exumi}
Yue Xu, Litao Wei, Pengyu An, Qingyu Zhang, and Yong-Lu Li.
\newblock {exUMI}: Extensible robot teaching system with action-aware
  task-agnostic tactile representation.
\newblock In \emph{Proceedings of The 9th Conference on Robot Learning}, volume
  305 of \emph{Proceedings of Machine Learning Research}, pp.\  2536--2554.
  PMLR, 2025.

\bibitem[Yang et~al.(2022)Yang, Ma, Zhang, Zhu, Yuan, and Owens]{yang2022touch}
Fengyu Yang, Chenyang Ma, Jiacheng Zhang, Jing Zhu, Wenzhen Yuan, and Andrew
  Owens.
\newblock Touch and go: Learning from human-collected vision and touch.
\newblock In \emph{Thirty-Sixth Conference on Neural Information Processing
  Systems Datasets and Benchmarks Track}, 2022.

\bibitem[Yu et~al.(2025)Yu, Han, Wang, Saxena, Xu, and Zhao]{yu2025mimictouch}
Kelin Yu, Yunhai Han, Qixian Wang, Vaibhav Saxena, Danfei Xu, and Ye~Zhao.
\newblock {MimicTouch}: Leveraging multi-modal human tactile demonstrations for
  contact-rich manipulation.
\newblock In \emph{Proceedings of The 8th Conference on Robot Learning}, volume
  270 of \emph{Proceedings of Machine Learning Research}, pp.\  4844--4865.
  PMLR, 2025.

\bibitem[Yuan et~al.(2017)Yuan, Dong, and Adelson]{yuan2017gelsight}
Wenzhen Yuan, Siyuan Dong, and Edward Adelson.
\newblock {GelSight}: High-resolution robot tactile sensors for estimating
  geometry and force.
\newblock \emph{Sensors}, 17\penalty0 (12):\penalty0 2762, 2017.
\newblock \doi{10.3390/s17122762}.

\bibitem[Zhang et~al.(2018)Zhang, Isola, Efros, Shechtman, and
  Wang]{zhang2018perceptual}
Richard Zhang, Phillip Isola, Alexei~A Efros, Eli Shechtman, and Oliver Wang.
\newblock The unreasonable effectiveness of deep features as a perceptual
  metric.
\newblock In \emph{CVPR}, 2018.

\bibitem[Zhang et~al.(2026)Zhang, Zhou, Zhang, Desai, Amosa, Soleymanzadeh,
  Lei, Zhou, Zheng, and She]{zhang2026contactworld}
Zhiyuan Zhang, Pokuang Zhou, Kaidi Zhang, Adeesh Desai, Temitope Amosa, Davood
  Soleymanzadeh, Jiuzhou Lei, Yuhao Zhou, Minghui Zheng, and Yu~She.
\newblock {ContactWorld}: What representations matter for vision-tactile latent
  world models in contact-rich manipulation.
\newblock \emph{arXiv preprint arXiv:2606.13877}, 2026.

\bibitem[Zhao et~al.(2025)Zhao, Ma, Wang, and Adelson]{zhao2025t3}
Jialiang Zhao, Yuxiang Ma, Lirui Wang, and Edward Adelson.
\newblock Transferable tactile transformers for representation learning across
  diverse sensors and tasks.
\newblock In \emph{Proceedings of The 8th Conference on Robot Learning}, volume
  270 of \emph{Proceedings of Machine Learning Research}, pp.\  3766--3779.
  PMLR, 2025.

\bibitem[Zhen et~al.(2026)Zhen, Gao, Sun, Zhao, Yang, Du, Guo, Wang, Qiao, and
  Gan]{zhen2026actionimagesendtoendpolicy}
Haoyu Zhen, Zixian Gao, Qiao Sun, Yilin Zhao, Yuncong Yang, Yilun Du, Pengsheng
  Guo, Tsun-Hsuan Wang, Yi-Ling Qiao, and Chuang Gan.
\newblock Action images: End-to-end policy learning via multiview video
  generation, 2026.
\newblock URL \url{https://arxiv.org/abs/2604.06168}.

\bibitem[Zheng et~al.(2026)Zheng, Gu, Zheng, Li, Zang, Tian, Li, Hao, Gao, Liu,
  et~al.]{zheng2026omnivta}
Yuhang Zheng, Songen Gu, Yupeng Zheng, Weize Li, Yujie Zang, Shuai Tian, Xiang
  Li, Ce~Hao, Chen Gao, Si~Liu, et~al.
\newblock {OmniVTA}: Visuo-tactile world modeling for contact-rich robotic
  manipulation.
\newblock \emph{arXiv preprint arXiv:2603.19201}, 2026.

\bibitem[Zhou et~al.(2025)Zhou, Pan, LeCun, and Pinto]{zhou2025dinowm}
Gaoyue Zhou, Hengkai Pan, Yann LeCun, and Lerrel Pinto.
\newblock {DINO-WM}: World models on pre-trained visual features enable
  zero-shot planning.
\newblock In \emph{Proceedings of the 42nd International Conference on Machine
  Learning}, volume 267 of \emph{Proceedings of Machine Learning Research},
  pp.\  79115--79135. PMLR, 2025.

\bibitem[Zhou et~al.(2024)Zhou, Du, Chen, Li, Yeung, and
  Gan]{zhou2024robodreamer}
Siyuan Zhou, Yilun Du, Jiaben Chen, Yandong Li, Dit-Yan Yeung, and Chuang Gan.
\newblock {RoboDreamer}: Learning compositional world models for robot
  imagination.
\newblock In \emph{Proceedings of the 41st International Conference on Machine
  Learning}, volume 235 of \emph{Proceedings of Machine Learning Research},
  pp.\  61885--61896. PMLR, 2024.

\bibitem[Zhou et~al.(2019)Zhou, Barnes, Lu, Yang, and Li]{zhou2019continuity}
Yi~Zhou, Connelly Barnes, Jingwan Lu, Jimei Yang, and Hao Li.
\newblock On the continuity of rotation representations in neural networks.
\newblock In \emph{Proceedings of the IEEE/CVF Conference on Computer Vision
  and Pattern Recognition (CVPR)}, 2019.

\bibitem[Zhu et~al.(2025)Zhu, Wu, Guo, Liu, Cheang, and Kong]{zhu2025irasim}
Fangqi Zhu, Hongtao Wu, Song Guo, Yuxiao Liu, Chilam Cheang, and Tao Kong.
\newblock {IRASim}: A fine-grained world model for robot manipulation.
\newblock In \emph{Proceedings of the IEEE/CVF International Conference on
  Computer Vision (ICCV)}, 2025.
\newblock URL \url{https://arxiv.org/abs/2406.14540}.

\end{thebibliography}
\bibliographystyle{iclr2027_conference}

\appendix
\section{Dataset and Force-Estimation Details}
\label{app:dataset}

\begin{figure}[t]
    \centering
    \includegraphics[width=0.8\columnwidth]{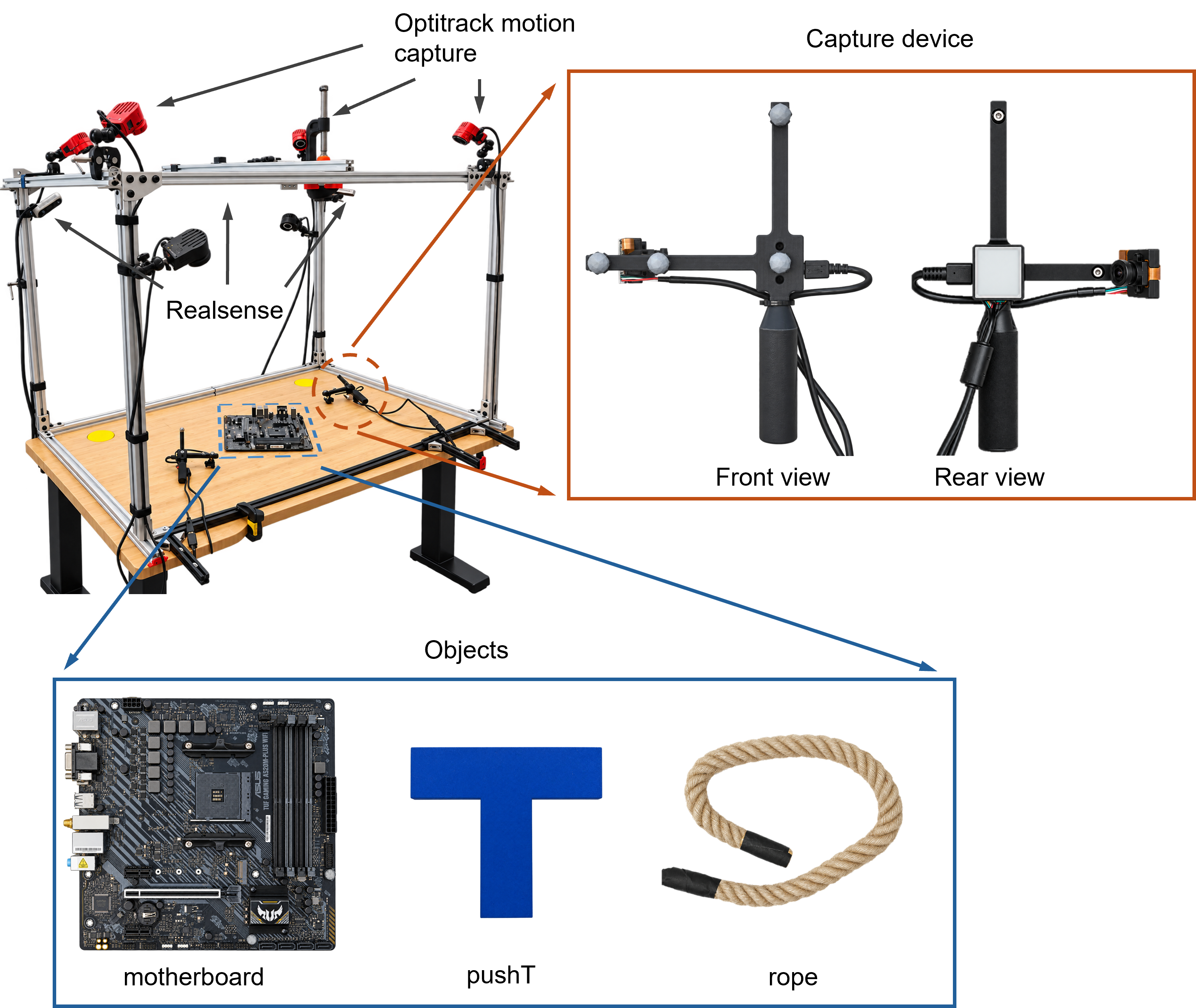}
    \caption{\textbf{Human data collection setup.} Left: the workspace and three fixed cameras. Top right: the two handheld devices and their sensing hardware. Bottom: the motherboard, PushT block, and rope used for collection.}
    \label{fig:data_collection_setup}
\end{figure}

\subsection{Dataset Composition}

The curated dataset contains 529,839 aligned time steps from 60 source recordings. Removing intervals with invalid sensor or tracking data leaves 187 variable-length segments, totaling 4.91~hours at the nominal 30~Hz alignment rate. Multiple segments may come from the same source recording and therefore do not represent independent collection sessions. Table~\ref{tab:dataset_statistics} gives the distribution across objects.

\begin{table}[t]
    \centering
    \caption{Dataset statistics after curation. Duration is computed from the number of aligned time steps at 30~Hz and counts each bimanual interaction once.}
    \label{tab:dataset_statistics}
    \begin{tabular}{lrrr}
        \toprule
        Object & Recordings & Segments & Duration (min) \\
        \midrule
        Motherboard & 12 & 25 & 93.9 \\
        PushT       & 30 & 92 & 96.4 \\
        Rope        & 18 & 70 & 104.1 \\
        \midrule
        Total       & 60 & 187 & 294.4 \\
        \bottomrule
    \end{tabular}
\end{table}

\subsection{Temporal Alignment and Curation}

Visual streams are recorded at a nominal 30~Hz and OptiTrack poses at 100~Hz. We align all modalities by timestamp to a common 30~Hz timeline. Because the tactile cameras operate at a lower frame rate, adjacent time steps may reuse the same tactile image. For each sensor, a freshness flag and source-frame index distinguish new captures from repeated observations.

We remove intervals containing camera freezes or invalid pose tracking and divide each recording into contiguous valid segments. Force is estimated for each new tactile capture and held constant on aligned rows that reuse the same image. We retain the source-frame indices to audit this alignment.

\subsection{Normal-Force Estimation}\label{sec:normal_force_estimation}

We calibrate a sensor-specific normal-force estimator for the GelSight Mini~\citep{yuan2017gelsight} sensors used in our collection setup. For each recording, we construct a reference image from low-contact observations and compute features describing the contact area, surface deformation, and RGB change relative to this reference.

We standardize these features and map them to normal force using support vector regression with a radial basis function kernel (RBF-SVR). The kernel provides a nonlinear mapping in which tactile observations with similar feature vectors receive similar force predictions. We train the regressor on separate CNC calibration data collected with known normal loads over 0.07--15.00~N. A reference-dependent noise gate suppresses responses near the no-contact state.

Position-grouped cross-validation on 4,709 CNC calibration samples gives a mean absolute error of 0.318~N (95\% CI: 0.300--0.338~N) and a mean absolute percentage error of 6.40\%. The percentage error is computed per sample relative to its measured force.

This calibration is specific to our sensors. Optical tactile images vary with the sensor unit, gel, illumination, camera response, and gel condition, so we do not assume that the mapping transfers unchanged to another sensor. The human interaction recordings also lack synchronized force ground truth. We use the estimates as approximate loading signals for action construction, not as measured force annotations.

\section{Evaluation Details}\label{app:evaluation_details}

\noindent\textbf{Prediction Protocol.}
Each test window contains eight observed frames followed by
eight target frames, corresponding to a prediction horizon
of approximately 0.53\,s at 15\,fps.
Models receive the observed frames and recorded action sequence.
We score only the predicted frames, separately for the scene
camera, left and right wrist cameras, and left and right
GelSight sensors~\citep{yuan2017gelsight}.
All checkpoints and ablations are evaluated on identical
test windows with identical random seeds.

\noindent\textbf{Reference Frames.}
For both human recordings and robot episodes, reference frames
are obtained by decoding ground-truth latents with the frozen RGB
and tactile VAEs.
These scores therefore measure prediction fidelity relative to
VAE reconstructions rather than raw recordings.

\noindent\textbf{Visuo-Tactile Prediction Metrics.}
We compute PSNR, SSIM~\citep{1284395}, and LPIPS~\citep{zhang2018perceptual} independently for each stream.
For a predicted window $\widehat{\mathbf{Y}}$ and its reference
$\mathbf{Y}$, PSNR is
\begin{equation}
    \operatorname{PSNR}
    = 10\log_{10}
    \frac{R^2}{\operatorname{MSE}(\widehat{\mathbf{Y}},\mathbf{Y})},
\end{equation}
where $R$ is the peak value of the image intensity range and
the mean squared error is computed over all pixels, channels,
and predicted frames in the window.
PSNR measures pixel-level fidelity and is sensitive to spatial
alignment and contact imprints, making it our primary metric
for the low-texture tactile streams.

SSIM~\citep{1284395} compares local luminance, contrast, and
structure. We compute it per predicted frame and average
over the window.
LPIPS~\citep{zhang2018perceptual} measures perceptual distance
using pretrained AlexNet~\citep{Krizhevsky2012ImageNet} features and is likewise averaged
over predicted frames.
Higher PSNR and SSIM and lower LPIPS indicate better fidelity.

\noindent\textbf{Zero-Shot Robot Evaluation.}
Models trained on human recordings are applied to held-out
robot episodes (Sec.~\ref{robot_data}) without fine-tuning.
Before inference, robot observations are mapped to the human
recording conventions for tactile color and illumination
and wrist-camera tone.
Action representations are constructed from OptiTrack sensor
poses and GelSight-estimated force using the procedure in
Sec.~\ref{sec:action_conditioning}.
We use OptiTrack rather than the robot's joint readings so that
human and robot actions are computed by the same pipeline from
the same tracking system.
The two sources disagree at the scale of a single step.
Per-step translations computed from forward kinematics differ
from those computed from OptiTrack by a median of 0.33\,mm
(90th percentile 0.99\,mm), while the sensor moves a median of
0.60\,mm per step (90th percentile 3.06\,mm).
Grouped by speed, the difference is 20--40\% of the per-step
motion, and it exceeds the motion when the robot is nearly still.
Per-step rotations differ by a median of 0.1--0.2$^\circ$.
Likely sources are the calibration between the robot base and the
OptiTrack frame, joint backlash, and arm compliance under contact
load, none of which forward kinematics observes.
Actions built from forward kinematics would therefore depart from
the human training data by a large fraction of each step, whereas
OptiTrack measures the sensor directly in both settings.
We select evaluation windows in which contact is made or broken
and apply the same eight-frame observation and eight-frame
prediction protocol.
All metrics are computed per stream against the VAE-reconstructed
reference frames described above.

\section{Qualitative Robot Rollouts}
\label{app:robot_qual}

\begin{figure}[t]
    \centering
    \includegraphics[width=1\columnwidth]{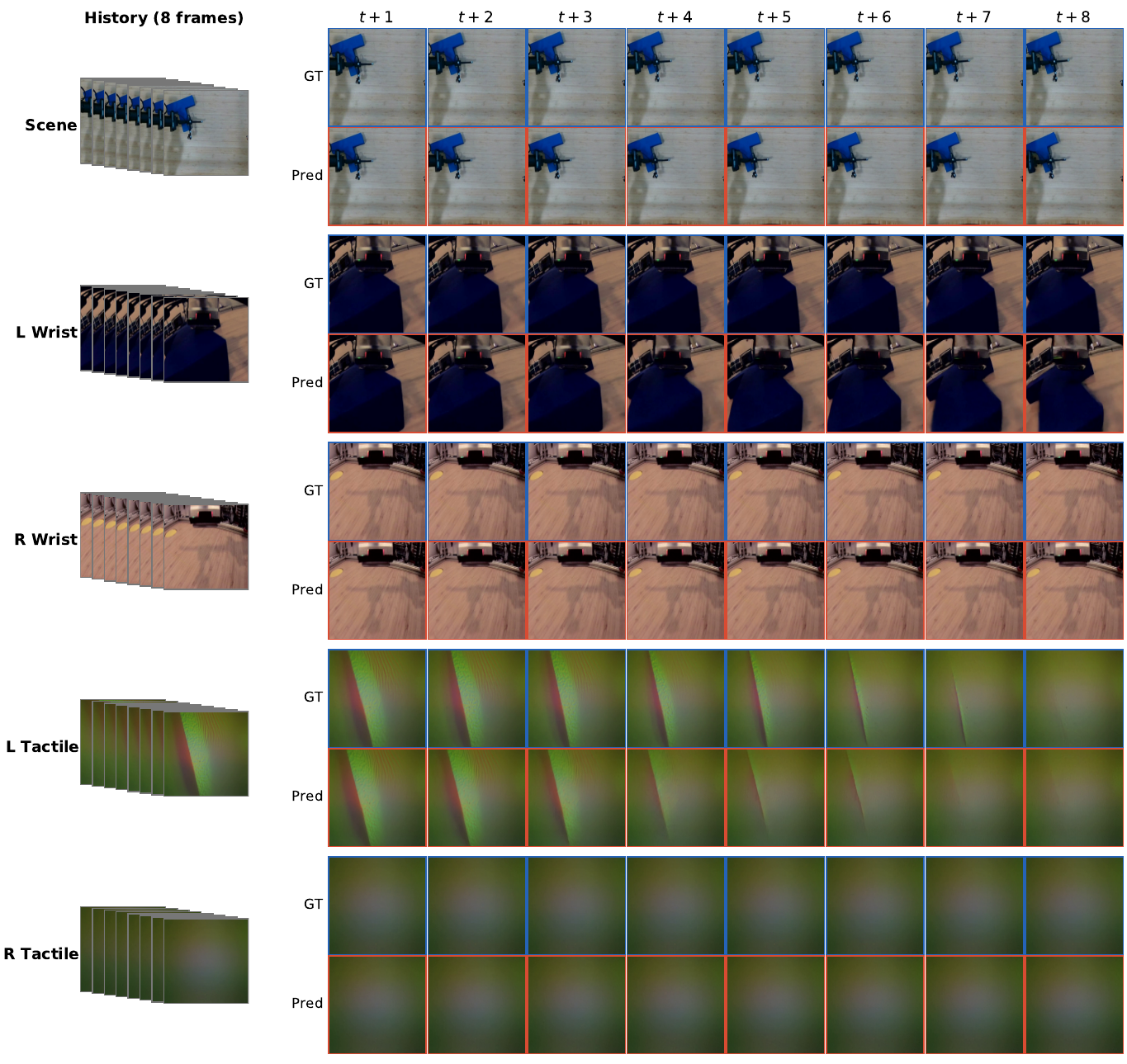}
    \caption{\textbf{Zero-shot rollout on robot teleoperation.} A held-out
    robot PushT window in which the sensor breaks contact with the block,
    predicted without fine-tuning by
    the model trained on human data. Given eight observed frames (left) and
    the robot's sensor actions, ACTS predicts the next eight frames of the
    scene view, both wrist views, and both GelSight streams. Ground truth in
    blue boxes, predictions in orange boxes. The scene and left tactile
    predictions follow the recording as the sensor lifts off. In the left
    wrist view, the recorded frames barely change, while the predicted block
    face grows and shifts from $t{+}4$ onward. More robot rollouts:
    \url{https://acts-anon.github.io/}.}
    \label{fig:robot_qual}
\end{figure}

Fig.~\ref{fig:robot_qual} shows a zero-shot rollout on a held-out robot
episode from the evaluation set of Sec.~\ref{robot_data}, produced by the
model trained only on human data and discussed in
Sec.~\ref{sec:robot_transfer}. Rollouts on all three objects are on the
project page.

\section{Tactile Encoder}\label{sec:tactile_encoder}

We adapt the pretrained causal 3D VAE of Wan2.2-TI2V-5B~\citep{wan2025} to tactile video with LoRA, keeping all base weights frozen. Training uses 6,533 left and right gel-sensor videos (64 held out for validation) from visuo-tactile datasets including
GelSLAM~\citep{huang2025gelslam},
FeelAnyForce~\citep{shahidzadeh2024feelanyforce},
Tactile MNIST~\citep{schneider2025tactilemnist},
FoTa~\citep{zhao2025t3},
TacQuad~\citep{feng2025learning},
3D Cal~\citep{kota2026threedcal},
and TactileTracking~\citep{huang2025normalflow}, covering 26 tasks that include manipulation, indentation, and object tracking. Each sample is a 17-frame clip resampled from 30 to 6 fps and center-cropped to 256$\times$256. Rank-32 LoRA~\citep{hu2021loralowrankadaptationlarge} adapters ($\alpha=32$) go on all 72 convolutional layers of the encoder and decoder, 30.6M trainable parameters (4.2\% of the VAE). The objective is an L1 reconstruction loss weighted by optical-flow magnitude, so regions of the gel that deform get more weight. The flow is AllTracker~\citep{harley2025alltrackerefficientdensepoint} reference flow from frame 0 to frame $t$. Two more terms are added: a spatial-gradient (edge) L1 loss (weight 1.0) and LPIPS (weight 0.1). We train for 4,000 optimizer steps with AdamW~\citep{loshchilov2019decoupledweightdecayregularization}, learning rate $10^{-4}$ with 100 warmup steps, an effective batch size of 4 and bf16, on a single H100. Validation PSNR is 40.2 dB.

\section{Action Representation Details}
\label{app:action_representation}

\noindent\textbf{Sensor-frame motion and force.}
For each hand $h\in\{L,R\}$, the sensor pose comprises
an orientation $R_t^h\in\mathrm{SO}(3)$ and position
$\mathbf{p}_t^h\in\mathbb{R}^{3}$ in a common reference frame.
The estimated normal force $f_t^h$ is obtained using the
procedure in Appendix~\ref{sec:normal_force_estimation}.

Sensor pose captures motion but does not fully describe
contact loading.
Following the force-informed formulation of
\citet{chen2025dexforceextractingforceinformedactions},
we interpret normal force as a virtual displacement along
the sensor's local pressing direction:
\begin{equation}
    \widetilde{\mathbf{p}}_t^h
    =
    \mathbf{p}_t^h
    + k_f f_t^h R_t^h\mathbf{n}_s,
    \label{eq:force_informed_position}
\end{equation}
where $\mathbf{n}_s$ is the unit gel normal in sensor
coordinates and $k_f=0.5\,\mathrm{mm}/\mathrm{N}$
is an assumed controller compliance.
The orientation remains unchanged.
As illustrated in Fig.~\ref{fig:teaser}a,
the offset represents a virtual controller displacement,
not a measurement of gel compression.

To construct actions, we express relative motion in the
sensor coordinate system at frame $t$ and augment its
translation with the change in force-induced displacement:
\begin{equation}
\begin{aligned}
    \Delta R_t^h
    &= (R_t^h)^\top R_{t+1}^h, \\
    \Delta\mathbf{p}_t^h
    &= (R_t^h)^\top
       (\mathbf{p}_{t+1}^h-\mathbf{p}_t^h), \\
    \Delta\widetilde{\mathbf{p}}_t^h
    &= \Delta\mathbf{p}_t^h
       + k_f\Delta f_t^h\mathbf{n}_s,
\end{aligned}
\end{equation}
where $\Delta f_t^h=f_{t+1}^h-f_t^h$.
The force increment is applied directly along the local
normal, rather than obtained by differencing the virtual
positions in Eq.~\ref{eq:force_informed_position}.
Concatenating the augmented translation with the
six-dimensional rotation representation of \citet{zhou2019continuity} gives
\begin{equation}
    \mathbf{a}_{t+1}^h
    =
    [\Delta\widetilde{\mathbf{p}}_t^h;
     \operatorname{Rot6D}(\Delta R_t^h)]
    \in\mathbb{R}^{9}.
\end{equation}
The index $t+1$ identifies the transition endpoint.

\noindent\textbf{Reference-point projection.}
We define three fixed points in sensor coordinates:
\begin{equation}
    \mathbf{o}_1=\mathbf{o}_c,
    \qquad
    \mathbf{o}_2=\mathbf{o}_c+\ell_n\mathbf{n}_s,
    \qquad
    \mathbf{o}_3=\mathbf{o}_c+\ell_p\mathbf{e}_s,
\end{equation}
where $\mathbf{o}_c$ is the gel center,
$\mathbf{e}_s$ is a unit in-plane axis, and
$\ell_n$ and $\ell_p$ are fixed offset distances.
These points jointly encode sensor position and orientation.

Their coordinates in the common reference frame are
$\mathbf{q}_{t,j}^h=R_t^h\mathbf{o}_j+\mathbf{p}_t^h$.
Using the calibrated transformation $(R_C,\mathbf{t}_C)$
from this frame to the scene camera, we obtain
\begin{equation}
    \mathbf{q}_{t,j}^{C,h}
    =R_C\mathbf{q}_{t,j}^h+\mathbf{t}_C,
    \qquad
    z_{t,j}^h=[\mathbf{q}_{t,j}^{C,h}]_3.
\end{equation}
We project each point using the scene-camera calibration
and normalize its pixel coordinates to obtain
$(\bar{x}_{t,j}^h,\bar{y}_{t,j}^h)$.
Concatenating these coordinates and camera-space depths
for all three points produces $\mathbf{u}_t^h\in\mathbb{R}^{9}$.
Unlike $\mathbf{a}_{t+1}^h$, which describes a transition,
$\mathbf{u}_t^h$ describes sensor geometry at frame $t$.

\noindent\textbf{Action construction and normalization.}
Sensor and view actions each have nine dimensions per hand.
We use $\mathbf{n}_s=(0,-1,0)^\top$ and
$k_f=5\times10^{-4}\,\mathrm{m/N}$. Rot6D concatenates the first
two columns of the relative rotation matrix. View conditions
contain $(x/W,y/H,d)$ for each of the three reference points,
with depth $d$ in meters. Sensor and view representations are
normalized separately, pooling both hands to estimate training-set
percentiles:
\begin{equation}
    \widetilde b_j
       =2\frac{b_j-q_{0.01,j}}{q_{0.99,j}-q_{0.01,j}}-1.
\end{equation}

\section{Implementation and Training Details}
\label{sec:implementation_details}

\subsection{Data Preparation}
\label{sec:train_details}

\noindent\textbf{Dataset splits.}
We train separate models for PushT, Rope, and Motherboard
using the same architecture and training recipe.
Each dataset is partitioned by complete source episodes,
preventing windows from the same episode from appearing
in different splits.
Table~\ref{tab:implementation_splits} reports episode counts,
window counts, and optimization steps per epoch.
The PushT training set excludes windows with invalid action
transitions.
For Motherboard, two test episodes are in-distribution,
while three were recorded in a separate session with different
hardware and force calibration and are treated as
out-of-distribution.
Wrist observations are available in every episode.

\begin{table*}[t]
\centering
\footnotesize
\setlength{\tabcolsep}{3pt}
\renewcommand{\arraystretch}{1.08}
\caption{Dataset splits and training sizes.
Splits are defined by complete source episodes.
Window counts correspond to examples used for training
and evaluation.}
\label{tab:implementation_splits}
\vspace{1ex}
\begin{tabular*}{\textwidth}{
    @{\extracolsep{\fill}}
    lrrrrrrr
    @{}
}
\toprule
\multirow{2}{*}{Task}
& \multicolumn{3}{c}{Episodes}
& \multicolumn{3}{c}{Windows}
& \multirow{2}{*}{Steps/epoch} \\
\cmidrule(lr){2-4}\cmidrule(lr){5-7}
& Train & Val & Test
& Train & Val & Test & \\
\midrule
PushT
& 56 & 4 & 8
& 55,199 & 4,439 & 5,756 & 6,900 \\
Rope
& 53 & 8 & 9
& 77,188 & 6,363 & 8,468 & 9,648 \\
Motherboard
& 21 & 2 & 5
& 71,386 & 5,411 & 16,699 & 8,923 \\
\bottomrule
\end{tabular*}
\end{table*}

\noindent\textbf{Observation preprocessing.}
We sample all five observation streams at 15\,fps and
resize and center-crop frames to $256\times256$.
Each window contains eight history frames and eight
future frames.
The scene and two wrist streams use the pretrained
Wan2.2-TI2V-5B VAE~\citep{wan2025}, while both tactile streams use its
GelSight-specific~\citep{yuan2017gelsight} LoRA-adapted variant.
All VAEs remain frozen, and observation latents are
encoded offline and cached.
The VAEs produce 48-channel latents with temporal and
spatial compression factors of four and sixteen.
We prepend a copy of the initial frame before encoding
and remove the resulting anchor latent, obtaining
$48\times4\times16\times16$ latents per stream.
The four latent blocks comprise two history blocks
and two future blocks.
Spatial patches of size $2\times2$ yield an $8\times8$
token grid per block, with 192 input features per token.

\subsection{Model Initialization and Adaptation}

\noindent\textbf{Backbone.}
We initialize the generator branch of
Cosmos3-Nano~\citep{nvidia2026cosmos3omnimodalworld}
from the official checkpoint.
It contains 36 transformer layers with hidden dimension
4,096. The understanding branch is not used.
We retain the pretrained generator, RGB input and output
projections, final normalization layers, and noise-level
embedder.
The native action projection is replaced by our action
interfaces, initialized from its pretrained weights.
Each task starts independently from the same checkpoint
with fresh adapters and interfaces, without warm-starting
from another task's trained model.

\noindent\textbf{LoRA adapters.}
The base generator remains frozen.
We apply LoRA~\citep{hu2021loralowrankadaptationlarge}
with rank $16$ and scaling parameter $\alpha=32$ to
the query, key, value, output, gate, up, and down projections
in every transformer layer.
These $36\times7=252$ linear layers introduce 43.6M
trainable parameters.
The LoRA matrices $A$ and $B$ are initialized with
Kaiming-uniform values and zeros, respectively, yielding
zero initial adapter updates.

\noindent\textbf{Modality interfaces.}
The scene and wrist streams share the frozen RGB
input and output projections.
Both tactile streams share a trainable projection pair
initialized from the corresponding RGB projections.
Sensor and view actions use separate nine-dimensional
input projections, each shared by both hands.
These projections are initialized from the first nine
input channels of the UMI-domain slice of the pretrained
action-conditioning projection.
All input projections map to the same hidden dimension.

\noindent\textbf{Stream embeddings and parameter count.}
A learned $9\times4096$ embedding distinguishes five
observation streams and four action streams.
The embeddings are initialized from a truncated normal
distribution, with action-stream rows additionally offset
by the pretrained action-modality embedding.
Together, the LoRA~\citep{hu2021loralowrankadaptationlarge} adapters, tactile and action projections,
and stream embeddings contain 45.3M trainable parameters,
approximately 0.30\% of the 15.2B backbone.
The RGB projections and noise-level embedder remain frozen.

\subsection{Sequence Packing and Temporal Alignment}
\label{app:acts_packing}

\noindent\textbf{Token layout.}
We concatenate streams into contiguous slices, ordering
observation tokens by latent block, patch row, and patch column.
A complete window contains
$5\times4\times64=1{,}280$ observation tokens,
$2\times15=30$ sensor-action tokens, and
$2\times16=32$ view-action tokens, totaling 1,342 tokens.
Each token retains its stream identity, frame index,
spatial coordinates, and target indicator.
Positional coordinates and attention masks are derived
from this metadata rather than storage offsets.

\noindent\textbf{Temporal and spatial coordinates.}
\label{app:acts_mrope}
All streams use a shared 15\,fps clock.
Observation blocks receive the endpoint indices
$3,7,11,15$ of their four-frame intervals.
Sensor-action tokens receive transition endpoints
$1,\ldots,15$, and view-action tokens receive frame
indices $0,\ldots,15$.
We convert frame index $k$ to the pretrained temporal scale:
\begin{equation}
    \tau(k)=r_{\mathrm{C}}\frac{k}{r},
    \qquad r=15,\quad r_{\mathrm{C}}=6.
    \label{eq:acts_temporal_coordinate}
\end{equation}
Observation tokens use patch row and column indices
as spatial coordinates.
Both spatial coordinates are zero for action tokens.
Frame indices retain their original values across
prediction prefixes within a window, so extending
the prefix does not change existing token coordinates.

\subsection{Training Procedure}

\noindent\textbf{Teacher-forcing prefixes.}
For each future block $q$, we construct a separate
prediction prefix with clean context
\begin{equation}
    \mathcal{C}_q
    = \left(Z_H,Z_{F,<q},\mathcal{A}\right),
    \label{eq:acts_training_context}
\end{equation}
where $Z_H=\{Z_H^m\}_{m\in\mathcal{M}}$ contains observation
histories,
$Z_{F,<q}=\{(z_1^m,\ldots,z_{q-1}^m)\}_{m\in\mathcal{M}}$
contains preceding ground-truth future blocks, and
$\mathcal{A}$ contains both action representations for
both hands over the complete window.
The current target appears only in noisy form.
Later observation blocks and clean copies of the current
target are excluded to prevent ground-truth leakage.
We apply the attention mask in
Sec.~\ref{sec:multimodal_transformer}, retaining access
to the full action trajectory.

Each window produces two teacher-forced passes.
The first contains two history blocks and the first
future block from each stream, while the second additionally
includes the preceding future block as clean context.
With all 62 action tokens retained, these prefixes contain
1,022 and 1,342 tokens, respectively.
Each pass contains 320 current target tokens.
Their losses contribute equally to one optimizer update.

\noindent\textbf{Noise sampling.}
We sample one noise level $\sigma$ per window and share
it across both target blocks and all observation streams.
Following the pretrained generator's Waver rule with
shift parameter $3.0$, we sample
\begin{align}
    \xi &\sim \mathcal{U}(0,1), \\
    \eta
    &= 1-\xi-1.29\left[
       \cos^2\!\left(\frac{\pi\xi}{2}\right)-1+\xi
       \right], \\
    \sigma &= \frac{3\eta}{1+2\eta}.
\end{align}
Gaussian noise $\epsilon_q^m$ is sampled independently
for each block and stream, with identity covariance.
History, preceding ground-truth blocks, and actions
remain clean.
Noise-level embeddings are applied only to current
target observation tokens.

\noindent\textbf{Flow-matching supervision.}
The interpolation in Eq.~\ref{eq:flow_interpolation}
defines the target velocity
$v_q^{m,*}=\epsilon_q^m-z_q^m$.
In Eq.~\ref{eq:flow_loss}, each stream's squared error
is normalized by its latent-block size $D_z$.
Stream losses are summed with equal weight and averaged
over future blocks.
We apply no additional noise-dependent loss weighting,
and clean context tokens receive no prediction loss.
The expectation is over training windows and sampled noise.

\noindent\textbf{Optimization.}
We use
AdamW~\citep{loshchilov2019decoupledweightdecayregularization}
with learning rate $5\times10^{-5}$,
$(\beta_1,\beta_2)=(0.9,0.95)$, and weight decay $0.01$.
The learning rate warms up for 100 steps and remains
constant thereafter.
Gradients are clipped to a global norm of $1.0$.
Training uses four NVIDIA H100 GPUs with two windows
per GPU, giving a global batch size of eight.
Trainable parameters are maintained in FP32 with BF16
autocast, and activation checkpointing is enabled.
Each model is trained for 20 epochs, with a checkpoint
saved after every epoch.

\subsection{Inference and Rollout}

\noindent\textbf{Within-window generation.}
Inference follows the training prefix construction,
replacing preceding ground-truth future blocks with
generated latents.
For each target block, all streams are initialized with
independent Gaussian noise and jointly denoised using
UniPC with 35 steps, without classifier-free guidance.
The sampler integrates the learned velocity field from
noise to clean latents, corresponding to $\sigma=1$
to $\sigma=0$.
Only current target latents are updated.
Observed history, previously generated blocks, and
the complete action window remain fixed.
The first generated block serves directly as clean
latent context for the second, without intermediate
decoding or re-encoding.
After restoring the VAE anchor, the corresponding
decoders $\{\mathcal{D}_m\}_{m\in\mathcal{M}}$
reconstruct the eight predicted frames per stream.

\noindent\textbf{Across-window rollout.}
To extend generation beyond one window, we use the
eight predicted frames as the next observation history.
For each stream, we prepend a copy of its first predicted
frame and re-encode the sequence, obtaining a new anchor
and two history latent blocks.
We advance the action window by eight frames and reset
the temporal origin for the next prediction window.
Thus, blocks within a window share latent context directly,
while successive windows pass context through decoded
and re-encoded observations.

\section{VT-WM-style Baseline}
\label{app:vtwm}

\noindent\textbf{Predictor.}
A 12-layer transformer, hidden dimension 384, 6 heads, MLP ratio 4; 116.6\,M
parameters. Inputs are the cached Wan2.2 latents our model uses:
$48\times4\times16\times16$ per stream, five streams, $2\times2$ patches giving 64
tokens per latent block. Attention is factorized into spatial and temporal passes
with axial RoPE. The three camera streams share an input projection and output
head; the two tactile streams share a second pair. Training minimizes $L_1$ on
the next latent block, teacher-forced and sampled losses weighted equally, with
the horizon-4 rollout generated without gradients.

\noindent\textbf{Action conditioning.}
After every self-attention block, camera and tactile tokens cross-attend to
action tokens, which carry sinusoidal positional embeddings. Actions are the
force-informed sensor deltas our model receives, 36 dimensions per frame. One
latent transition spans five frames; each receives its five raw transitions
concatenated (180 dimensions) rather than averaged, since averaging shares raw
transitions 3, 7 and 11 between adjacent latent transitions and leaks a later
action into an earlier prediction. Latent block $t$ is conditioned on the action
producing block $t{+}1$, the block it is supervised against.

\noindent\textbf{Training.}
AdamW, peak learning rate $3\times10^{-4}$, $(\beta_1,\beta_2)=(0.9,0.95)$,
weight decay 0.01, gradient clipping at 1.0, 500 warmup steps into cosine decay.
Batch size 8 with accumulation 4, bfloat16, one A100-80GB, 10 epochs. Early
stopping is off so every task gets the same budget.

\noindent\textbf{Deviations.}
(i) The cross-attention residual is gated by a zero-initialized scalar; the
backbone relies on adaLN-zero identity at initialization, which an ungated
sub-layer breaks.
(ii) Context is eight frames at 15\,fps (0.53\,s) rather than nine at 6\,fps
(1.5\,s), inherited from our encoder.
(iii) Tactile tokens come from our GelSight tokenizer rather than
Sparsh-X~\citep{higuera2025tactile} on Digit~360~\citep{lambeta2024digitizing}, which needs
audio, IMU and pressure channels this dataset lacks.
(iv) Three camera views rather than one, matching our model's inputs.
(v) 116.6\,M parameters for 20 epochs on one GPU, against 173\,M for 80k updates
on 32 A100s.

\noindent\textbf{Validation.}
Before training we check identity at initialization, action reachability,
temporal causality and gate-gradient flow. During training we track \emph{action
gain}: rollout $L_1$ under a held control action divided by the same under the
recorded action. A value near 1 means the model ignores its conditioning, which
validation loss cannot detect. The control is $\mathrm{Rot6D}(I)$ with zero
translation, not a zero vector: an all-zero rotation is degenerate and never
occurs in training. Our runs reach 1.47--1.87.

\noindent\textbf{Checkpoint selection.}
ACTS and VT-WM-style are evaluated at their final epoch. VT-WM-style's validation $L_1$ bottoms
at epoch 4--6, but we do not select on it: $L_1$ is minimized by the conditional
mean, and across the nine camera task--stream pairs the best-validation
checkpoint wins on PSNR and loses on LPIPS in eight. Action gain also rises
monotonically to the final epoch on all three tasks while rollout $L_1$ bottoms
mid-training. Selecting on validation instead moves scene-view PSNR by under
0.1\,dB on all three tasks and 0.4--0.7\,dB on the motherboard wrist streams, and
changes no conclusion.

\section{Force-Informed Action Ablation}
\label{app:force_ablation}

Tab.~\ref{tab:ablation_force} compares models trained with and without the
force term, on identical test windows. Both are trained for 10 epochs rather
than the 20 used for Tab.~\ref{tab:quantative_results}, which halves the cost
of the ablation. The 20-epoch models of Tab.~\ref{tab:quantative_results}
differ from the 10-epoch force-informed models by up to 0.6\,dB PSNR per
stream, so only the paired comparison within Tab.~\ref{tab:ablation_force} is
meaningful. Scene and wrist PSNR change by at most 0.3\,dB and never
significantly, as expected: the force term only alters the normal component
of the sensor action. Tactile PSNR differs by at most 0.6\,dB. The differences
are significant for both PushT tactile streams (PSNR), the left Motherboard
sensor (SSIM and LPIPS), and the right Rope sensor, where PSNR drops by
0.3\,dB.

\begin{figure}[t]
    \centering
    \includegraphics[width=1.0\columnwidth]{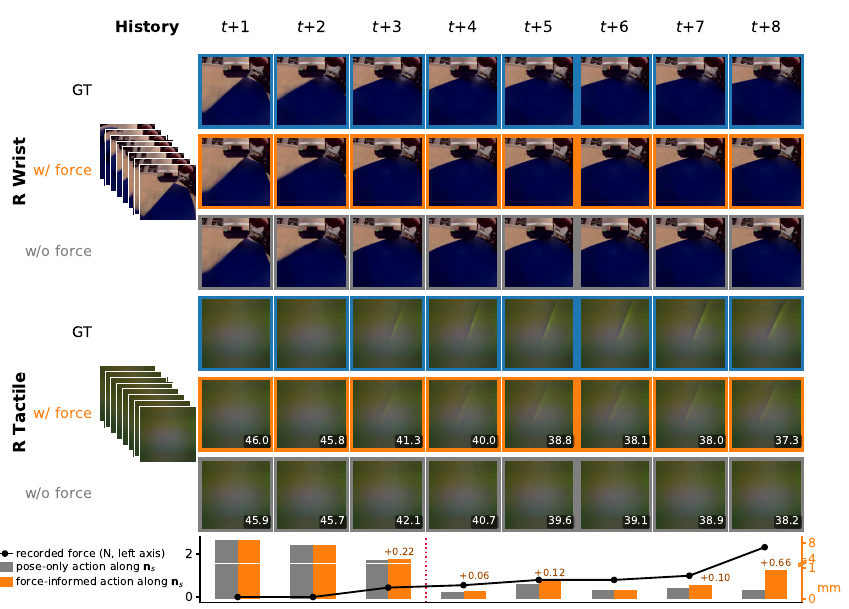}
    \caption{\textbf{Effect of the force term on tactile prediction.} A
    held-out PushT window in which the right sensor touches the block. Rows:
    ground truth (blue), force-informed actions (orange), pose-only actions
    (gray) for the right wrist and GelSight streams; numbers are per-frame
    PSNR. Bottom: recorded normal force (black, left axis) and per-step action
    along the gel normal $\mathbf n_s$ (Eq.~\ref{eq:sensor_action}; right axis,
    broken), pose-only (gray) and force-informed (orange), with the force
    increment $k_f\,\Delta f$ above each orange bar; the dotted line marks
    contact. The force term adds 1.2\,mm over the window, and only the model
    that receives it forms the imprint.}
    \label{fig:force_ablation_qual}
\end{figure}

These averages hide what the term does. Contact changes occupy few windows
and an imprint occupies few pixels, so frame metrics are dominated by static
gel. Fig.~\ref{fig:force_ablation_qual} shows a PushT window in which the
right sensor touches the block. The pose-only action advances the sensor
21.7\,mm along the gel normal over the eight predicted steps, 20.1\,mm of it
before contact; the force term adds 1.2\,mm in total, up to 0.66\,mm per step.
Yet only the force-conditioned model forms an imprint. The sensor-frame pose says
that the sensor advances, not where the block surface is; the force change
marks the contact. The pose-only prediction stays blank and still scores
slightly higher PSNR, since a blank gel costs less than a faint, slightly
displaced imprint. The converse case, a release in which the pose-only
model holds the imprint two frames past lift-off (43.8 vs.\ 37.0\,dB), is on
the project page: there the pose moves under 1\,mm along the normal while
the force-informed action retreats 1.9\,mm.

\begin{table}[t]
\centering
\footnotesize
\setlength{\tabcolsep}{3pt}
\renewcommand{\arraystretch}{1.08}
\caption{
Ablation of force information in action conditioning.
All observation streams are retained.
A checkmark indicates that force information is included
in the action representation.
Within each dataset, bold values indicate a significant difference
between the force and no-force configurations for the same
observation stream (paired $t$-test, $p<0.05$).
All configurations are evaluated after 10 training epochs
for a fair comparison.
}
\label{tab:ablation_force}
\vspace{1ex}
\begin{tabular*}{\columnwidth}{
    @{\extracolsep{\fill}}
    lcl
    S[table-format=2.2]
    S[table-format=1.4]
    S[table-format=1.4]
    @{}
}
\toprule
Task & Force & Stream
& {PSNR$\uparrow$}
& {SSIM$\uparrow$}
& {LPIPS$\downarrow$} \\
\midrule

\multirow{10}{*}{Rope}
& \multirow{5}{*}{--}
& Video         & 28.50 & 0.9021 & 0.0495 \\
& & Wrist (L)   & 21.68 & 0.7583 & 0.1444 \\
& & Wrist (R)   & 23.62 & 0.7966 & 0.1353 \\
& & Tactile (L) & 45.90 & 0.9874 & 0.0230 \\
& & Tactile (R) & {\textbf{46.30}} & 0.9898 & 0.0188 \\
\cmidrule(lr){2-6}
& \multirow{5}{*}{$\checkmark$}
& Video         & 28.56 & 0.9028 & 0.0490 \\
& & Wrist (L)   & 21.84 & 0.7639 & 0.1401 \\
& & Wrist (R)   & 23.45 & 0.7932 & 0.1354 \\
& & Tactile (L) & 45.73 & 0.9876 & 0.0221 \\
& & Tactile (R) & {\textbf{46.00}} & 0.9893 & 0.0210 \\

\midrule
\multirow{10}{*}{PushT}
& \multirow{5}{*}{--}
& Video         & 29.25 & 0.9202 & 0.0408 \\
& & Wrist (L)   & 22.61 & 0.8286 & 0.1067 \\
& & Wrist (R)   & 22.62 & 0.8432 & 0.1102 \\
& & Tactile (L) & {\textbf{45.90}} & 0.9868 & 0.0256 \\
& & Tactile (R) & {\textbf{47.71}} & 0.9900 & 0.0148 \\
\cmidrule(lr){2-6}
& \multirow{5}{*}{$\checkmark$}
& Video         & 29.14 & 0.9199 & 0.0409 \\
& & Wrist (L)   & 22.29 & 0.8222 & 0.1110 \\
& & Wrist (R)   & 22.52 & 0.8412 & 0.1130 \\
& & Tactile (L) & {\textbf{46.52}} & 0.9875 & 0.0219 \\
& & Tactile (R) & {\textbf{47.91}} & 0.9900 & 0.0149 \\

\midrule
\multirow{10}{*}{Motherboard}
& \multirow{5}{*}{--}
& Video         & 27.79 & 0.8798 & 0.0468 \\
& & Wrist (L)   & 17.71 & 0.5280 & 0.2597 \\
& & Wrist (R)   & 20.73 & 0.6863 & 0.1626 \\
& & Tactile (L) & 41.09 & {\textbf{0.9778}} & {\textbf{0.0410}} \\
& & Tactile (R) & 45.53 & 0.9882 & 0.0273 \\
\cmidrule(lr){2-6}
& \multirow{5}{*}{$\checkmark$}
& Video         & 27.70 & 0.8783 & 0.0482 \\
& & Wrist (L)   & 17.57 & 0.5275 & 0.2598 \\
& & Wrist (R)   & 20.70 & 0.6842 & 0.1612 \\
& & Tactile (L) & 41.36 & {\textbf{0.9800}} & {\textbf{0.0360}} \\
& & Tactile (R) & 45.73 & 0.9886 & 0.0277 \\
\bottomrule
\end{tabular*}
\end{table}

\section{ContactWorld Baseline}
\label{sec:contactworld_baseline}

ContactWorld~\citep{zhang2026contactworld} is an action-conditioned
visuo-tactile world model based on JEPA, with modality-specific
CNN encoders and a GRU latent predictor but no image decoder.
Each model observes one visual and one tactile stream with
single-frame context.
After verifying our setup using its released simulation
checkpoints, we train ContactWorld from scratch using the
official code and hyperparameters for approximately 100k steps.
We use the same training episodes, source frames,
36-dimensional normalized actions, and held-out evaluation
windows as ACTS.

\noindent\textbf{Five-stream evaluation.}
To cover all five observation streams without modifying the
architecture, we train three models per task on
(scene, tactile-left), (wrist-left, tactile-left), and
(wrist-right, tactile-right).
Scene predictions come from the first model, left wrist and
tactile predictions from the second, and right wrist and
tactile predictions from the third.
Each model retains its original two-stream, single-frame
context, whereas ACTS conditions on all five streams over
eight frames.

\noindent\textbf{Pixel-space readout.}
We freeze the trained ContactWorld models and fit a separate
readout for each output stream to enable pixel-space evaluation.
Each readout maps four consecutive predicted ContactWorld
latents, each comprising 512-dimensional visual and tactile
embeddings, to one temporally compressed Wan2.2~\citep{wan2025} latent block.
A two-layer MLP embeds these latents and broadcasts its output
over the spatial latent grid.
Conditioned additionally on the last ground-truth history
VAE latent, four convolutional layers predict a residual
relative to that history latent.
Each readout contains approximately 1.5M parameters and uses
a zero-initialized output layer.
Readouts are trained on the GRU's own rollouts from training
windows using latent-space MSE and validation early stopping,
while ContactWorld remains frozen.
Predicted VAE latents are decoded and evaluated using the
same modality-specific VAEs and metrics as ACTS.

We additionally evaluate an oracle-input condition using
encoded ground-truth future frames and a context-only control
replacing ContactWorld latents with their training-set mean.
These conditions assess readout performance without latent
rollout error and without predictive latent information,
respectively.

\end{document}